\documentclass[runningheads]{llncs}
\usepackage{graphicx}

\usepackage{verbatim}

\usepackage{float}
\usepackage{subcaption}
\usepackage{adjustbox}
\usepackage[labelfont=bf,textfont=bf]{caption}
\usepackage[ruled,vlined,linesnumbered]{algorithm2e}
\usepackage{cite}
\usepackage{amsmath,amssymb,amsfonts}
\usepackage{textcomp}
\usepackage{xcolor}
\usepackage{todonotes}

\usepackage{array}
\usepackage{graphicx}
\usepackage{etoolbox}
\usepackage{rotating}

\usepackage{multicol}
\usepackage{multirow}

\usepackage[bottom]{footmisc}

\usepackage{orcidlink}

\begin{document}
\title{Evaluation of group fairness measures in student performance prediction problems}
\titlerunning{Evaluation of group fairness measures}
%
\author{Tai Le Quy\inst{1}\orcidlink{0000-0001-8512-5854} \and
Thi Huyen Nguyen\inst{1} \orcidlink{0000-0001-8195-716X} \and
Gunnar Friege\inst{2}\orcidlink{0000-0003-3878-9230}  \and
Eirini Ntoutsi\inst{3}\orcidlink{0000-0001-5729-1003} }
\authorrunning{T. Le Quy et al.}
%
\institute{L3S Research Center, Leibniz University Hannover, Hanover, Germany \\
\email{\{tai, nguyen\}@l3s.de}\\
\and
Institute for Didactics of Mathematics and Physics, Leibniz University Hannover, Hanover, Germany\\
\email{friege@idmp.uni-hannover.de}
\and 
Institute of Computer Science, Free University Berlin, Berlin, Germany\\
\email{eirini.ntoutsi@fu-berlin.de}}

\maketitle              
\begin{abstract}
Predicting students' academic performance is one of the key tasks of educational data mining (EDM). Traditionally, the high forecasting quality of such models was deemed critical. More recently, the issues of fairness and discrimination w.r.t. protected attributes, such as gender or race, have gained attention. Although there are several fairness-aware learning approaches in EDM, a comparative evaluation of these measures is still missing. In this paper, we evaluate different group fairness measures for student performance prediction problems on various educational datasets and fairness-aware learning models. Our study shows that the choice of the fairness measure is important, likewise for the choice of the grade threshold. 

 \and machine learning \and educational data mining
\end{abstract}
\section{Introduction}
\label{sec:introduction}

Educational data mining (EDM) applies data mining, artificial intelligence (AI), and machine learning (ML) to improve academic experiences. In recent years, AI-infused technologies have been widely studied and deployed by many educational institutions \cite{AJ2020preprocessing,hussain2018educational}. One of the most important tasks in EDM that attract great attention is student performance prediction. The early estimation of student learning outcomes can help detect and notify students at risk of academic failure. Besides, it supports institutional administrators in identifying key factors affecting students' grades and providing suitable interventions for outcome improvement. The performance prediction process relies on historical academic records and trains ML algorithms on labeled data to predict students' performance. Various datasets~\cite{Wightman1998, cortez2008using, kuzilek2017open} and approaches~\cite{Bindhia2019,Anupam2021,Lubna2019} have been proposed for the purpose. With the widespread use and benefits of AI systems, fairness has become a crucial criterion in designing such systems.     
 
Non-discriminative ML models have been a topic of increasing importance and growing momentum in education. Despite advances and superior accuracy of recent ML models, some studies have shown that ML-based decisions can be biased to protected attributes such as gender or race due to historical discrimination embedded in the data~\cite{mehrabi2021survey,ntoutsi2020bias}. Endeavoring to reduce biases is important and decisive in the applicability of an ML model in education. As an example, a recent study has proposed approaches that aim at predicting calculated grades of students in England as a replacement for actual grades due to the cancellation of exams during COVID-19~\cite{Anders2020}. However, the proposal could not be applied as a consequence of some exposed historical biases.  
 
 A large variety of fairness measures have been introduced in ML area. However, choosing proper measures can be cumbersome due to the dependence of fairness on context. There are more than 20 different fairness measures introduced in the computer science research area \cite{verma2018fairness,mehrabi2021survey}. In fact, no metric is universal and fits all circumstances \cite{foster2016big,verma2018fairness,mehrabi2021survey}. 
 Model developers should explore various fairness measures to decide the most appropriate notions for the context. Fairness is a fundamental concept of education, whereby all students must have an equal opportunity in study or be treated fairly regardless of their household income, assets, gender, or race \cite{meyer2014education}. Fairness definitions in education, hiring, and ML in the 50-year history have been discussed in the research of \cite{hutchinson201950}. However, no previous work exists on the efficiency of different fairness metrics and how to choose them in educational settings.

In this paper, we provide a comprehensive study to evaluate the sufficiency of various fairness metrics in student performance prediction. We consider a group of the most prevalent fairness notions in ML. Various experiments are conducted on diverse educational datasets and evaluated using different fairness metrics. Our experiments provides users a broad view of unfairness from diverse aspects in an educational context. Besides, the results also guide the selection of suitable fairness measures to evaluate students' grade predictive models. 
We believe our contributions are crucial to alleviate the burden of choosing fairness measures for consideration and motivate further studies to improve the accuracy and fairness of student performance prediction models.

The rest of the paper is organized as follows. In Section~\ref{sec:related}, we present some closely related work on fairness-aware ML and student performance prediction. Section~\ref{sec:fairness_measures} describes the most popular group fairness measures in ML. Next, we conduct quantitative evaluations of predictive models on educational datasets and discuss the choice of suitable fairness metrics in Section~\ref{sec:evaluation}. Finally, we conclude the paper in Section~\ref{sec:conclusion}.

\vspace{-7pt}
\section{Related work}
\label{sec:related}
\vspace{-5pt}
Extensive research efforts have been conducted to provide useful insights into students' performance analysis and prediction~\cite{xiao2022survey}. Various ML models were tested on different problem settings. Cortez et al.~\cite{cortez2008using} presented an early study to predict the grades of secondary students in Portuguese and Mathematics classes. 
Their results showed that good predictive accuracy could be achieved when previous school period grades are available. Similarly, Berhanu et al.~\cite{Fiseha2015} employed Decision Tree to predict students' performance using the agriculture college dataset. Some studies~\cite{Lubna2019,Nida2021} proposed diverse approaches to forecast students' grades in higher education. Besides, many other studies were reviewed in multiple surveys~\cite{Saleem2021,Amirah2015,Abdallah2020,Saleem2021,Saa2019}. They pointed out the most common techniques such as Decision Tree, Naive Bayes, Support Vector Machines, and neural networks and dominant factors impacting predictive outcomes (i.e., Cumulative Grade Point Average, previous grades, classroom attendance, etc.).

There are more than 20 fairness notions introduced for classification~\cite{verma2018fairness,mehrabi2021survey}. One of the most well-known fairness measure is \textit{demographic parity}, so-called \textit{statistical parity}. It requires an equal probability of positive predictions in protected and non-protected groups. However, Dwork et al.~\cite{dwork2012fairness} argued that the metric fails to ensure individual fairness. 
To avoid this, Hardt et al.~\cite{hardt2016equality} proposed \textit{equalized odds} metric. It measures whether a classifier predicts labels equally well for all values of attributes. Besides, many other popular metrics were introduced and used in fairness ML studies such as \textit{predictive parity}, \textit{predictive equality}~\cite{chouldechova2017fair}, \textit{treatment equality}~\cite{berk2018fairness}, etc. Despite a substantial number of fairness measures, there is no metric that fits all circumstances~\cite{mehrabi2021survey,verma2018fairness}.

Following the evolution of fairness measures, recent studies have attempted to evaluate fairness in an educational context \cite{gardner2019evaluating,Renzhe2020,Weijie2021}.
Anderson et al.~\cite{Anderson2019} conducted two post-hoc fairness assessments for existing student graduation prediction models.
Renzhe et al.~\cite{Renzhe2020} studied different combinations of student data sources for building highly predictive and fair models for predictions of college success.
Jiang et al. \cite{jiang2021towards} proposed several strategies to mitigate bias in the LSTM grade prediction model. They report experimental results on the true positive rate (TPR), true negative rate (TNR), and accuracy.

\vspace{-8pt}
\section{Fairness measures}
\label{sec:fairness_measures}
\vspace{-5pt}
\vspace{-20pt}
\begin{table}[!htb]
\centering
\caption{An overview of group fairness measures}\label{tbl:measures}
\begin{tabular}{lccr}
\hline
\multicolumn{1}{c}{\textbf{ Measures }} &  \multicolumn{1}{c}{\textbf{ Proposed by}} & \multicolumn{1}{c}{\textbf{  Published year  }} & \textbf{ \#Citations } \\ \hline
Statistical parity  & \cite{dwork2012fairness}&  2012 & 2,367 \\
Equal opportunity &  \cite{hardt2016equality}& 2016 & 2,575  \\
Equalized odds  & \cite{hardt2016equality} & 2016 & 2,575  \\
Predictive parity   & \cite{chouldechova2017fair} & 2017 & 1,430\\
Predictive equality &  \cite{corbett2017algorithmic}& 2017 & 878 \\
Treatment equality  & \cite{berk2018fairness}  & 2018 & 626 \\
Absolute Between-ROC Area  & \cite{gardner2019evaluating}&  2019 & 84 \\
\hline
\end{tabular}
\vspace{-20pt}
\end{table}

This section presents the most prevalent group fairness notions used in ML. The list of notions\footnote{The number of citations is reported by Google Scholar on $1^{st}$ August 2022.} is summarized in Table~\ref{tbl:measures}. To simplify, we consider the student performance prediction problem as a binary classification task, which is formalized as below:

Let $\mathcal{D}$ be a binary classification dataset with class attribute
$ Y = \{+, -\}$, e.g., $Y = \{pass, fail\}$. $S$ is a binary protected attribute, $S \in \{s,\overline{s}\}$, e.g., $S$ = ``gender'', $S \in \{female, male\}$. In which, 
\textit{s} is the discriminated group (\emph{protected group}), e.g., ``\textit{female}'', and $\overline{s}$ is the non-discriminated group (\emph{non-protected group}), e.g., ``\textit{male}''. The predicted outcome is denoted as $\hat{Y} = \{+, -\}$. The notions $s_{+}$ ($s_{-}$), $\overline{s}_{+}$ ($\overline{s}_{-}$) are used to denote the protected and non-protected groups for the positive (negative, respectively) class. 

We use a confusion matrix (Fig. \ref{fig:confusion_matrix}) to demonstrate the group fairness measures with an example of a dataset with 100 instances,  class $Y = \{pass, fail\}$. The protected attribute is ``gender'', and the protected group is ``female''; the distribution of ``female'':``male'' is 46:54. Examples of fairness measures in the following sub-sections are computed based on this confusion matrix.
\vspace{-25pt}
\newcommand\MyBox[2]{
  \fbox{\lower0.75cm
    \vbox to 1.3cm{\vfil
      \hbox to 3.3cm{\hfil\parbox{3.2cm}{#1\\#2}\hfil}
      \vfil}%
  }%
}
\begin{center}

\renewcommand\arraystretch{1.5}
\setlength\tabcolsep{0pt}
\begin{figure}[H]
\centering
\begin{tabular}{c >{\bfseries}r @{\hspace{0.6em}}c @{\hspace{0.4em}}c @{\hspace{0.5em}}l}
  \multirow{10}{*}{\rotatebox{90}{\parbox{2.5cm}{\bfseries\centering Actual class\\\phantom{} }}} & 
    & \multicolumn{2}{c}{\bfseries Predicted class} & \\
  & & \bfseries Positive +  & \bfseries Negative - &  \\
  & Positive + & \MyBox{\centering True Positive (TP)}{$TP_{prot} + TP_{non-prot}$ \textbf{70} (32:38) } & \MyBox{\centering False Negative (FN)}{$FN_{prot} + FN_{non-prot}$ \textbf{10} (4:6) } &  \\ [2.5em]
  & Negative - & \MyBox{\centering False Positive (FP)}{$FP_{prot} + FP_{non-prot}$ \textbf{9} (4:5)} & \MyBox{\centering True Negative (TN)}{$TN_{prot} + TN_{non-prot}$ \textbf{11} (6:5)} &  \\
\end{tabular}
\caption{The confusion matrix with an example}
\vspace{-20pt}
\label{fig:confusion_matrix}
\end{figure}
\vspace{-20pt}
\end{center}
\subsection{Statistical parity}
\label{subsec:Statistical_parity}

\emph{Statistical parity} (denoted as \emph{SP}) is a well-known group fairness measure \cite{dwork2012fairness}, whereby the output of any classifier satisfies statistical parity if the difference (bias) in the predicted outcome ($\hat{Y}$) between any two groups under study (i.e., $s$ and $\overline{s}$) is up to a predefined tolerance threshold $\epsilon$: 
\begin{equation}
\label{eqn:statistical_parity_def}
 P(\hat{Y}|S=s) - P(\hat{Y}|S=\overline{s}) \leq \epsilon.
\end{equation}
We use the violation of statistical parity~\cite{simoiu2017sp_benchmarking,zliobaite2015relation,lequy2022survey} to measure the bias of a classifier:
\begin{equation}
\label{eqn:statistical_parity}
SP =  P(\hat{Y}=+|S=\overline{s}) - P(\hat{Y}=+|S=s).
\end{equation}
The value range:  $SP \in [-1, 1]$,
with $SP=0$ indicating no discrimination, $SP\in (0,1]$ designating that the protected group is discriminated, and  $SP\in [-1,0)$ standing for \emph{reverse discrimination} (the non-protected group is discriminated). In our example (Fig.~\ref{fig:confusion_matrix}), this measure shows the proportion of ``\textit{pass}'' students between the two demographic subgroups. $SP = \dfrac{38+6}{54} - \dfrac{32+4}{46} \approx 0.0322$.

\subsection{Equal opportunity}
\label{subsec:Equal_opportunity}
\emph{Equal opportunity} (denoted as \emph{EO}) is proposed by Hardt et al. \cite{hardt2016equality}, whereby a binary predicted outcome $\hat{Y}$ satisfies equal opportunity w.r.t. the protected attribute $S$ and the class attribute $Y$ if:
\begin{equation}
P(\hat{Y} = + | S = s, Y = +) = P(\hat{Y} = + | S = \overline{s}, Y = +).    
\end{equation}
In other words, the protected and non-protected groups should have equal true positive rates (TPR)  \cite{mehrabi2021survey,verma2018fairness}, $TPR = \dfrac{TP}{TP + FN}$ (i.e., the classifier should give similar results for students of both genders with actual ``\textit{pass}'' class).  A classifier with equal false negative rates (FNR), $ FNR = \dfrac{FN}{TP + FN} $, will also have equal TPR \cite{verma2018fairness}. The equal opportunity can be measured by:
\begin{equation}
    EO =|P(\hat{Y} = - | Y = +, S = \overline{s}) - P(\hat{Y} = - | Y = +, S=s)|. 
\end{equation}
The value range: $EO \in [0,1]$; with 0 standing no discrimination and 1 indicating maximum discrimination. In our example, $EO = \mid\dfrac{38}{38+6}-\dfrac{32}{32+4}\mid \approx 0.0253$.

\subsection{Equalized odds}
\label{subsec:Equalized_odds}
A predictor $\hat{Y}$ is satisfied \emph{equalized odds} (denoted as \textit{EOd}) w.r.t. the
protected attribute $S$ and class label $Y$, if  ``$\hat{Y}$ and $S$ are independent conditional on $Y$'' ~\cite{hardt2016equality}.
Specifically, predicted true positive and false positive probabilities should be the same between male and female student groups.
\begin{equation}\label{eq: eq_odds_def}
   P(\hat{Y}=+|S=s,Y=y)= P(\hat{Y}=+|S=\overline{s},Y=y), \phantom{aaa} y\in \{+, -\}.
\end{equation}
Therefrom, we can measure the equalized odds as the following \cite{lequy2022survey,iosifidis2019adafair}: 
\begin{equation}
\label{eq:equalized_odds}
EOd = \sum_{y\in \{+,-\}}|P(\hat{Y}=+|S=s,Y=y) - P(\hat{Y}=+|S=\overline{s},Y=y)|.
\end{equation}

The value range: $EOd \in [0, 2]$; with 0 standing for no discrimination and 2 indicating the maximum discrimination. In our example, $EOd = |\dfrac{32}{32+4}-\dfrac{38}{38+6}| + |\dfrac{4}{4+6} - \dfrac{5}{5+5}| \approx 0.1253$.

\subsection{Predictive parity}
\label{subsec:Predictive_parity}
\emph{Predictive parity} \cite{chouldechova2017fair} (denoted as \emph{PP}) is satisfied if both protected and non-protected groups have an equal positive predictive value  (PPV) or \emph{Precision}, $PPV = \dfrac{TP}{TP + FP}$, i.e., the probability of a student predicted to ``\textit{pass}'' actually having ``\textit{pass}'' class should be the same, for both male and female students.
\begin{equation}
    P(Y = +|\hat{Y} = +, S = s) = P(Y = +|\hat{Y} = +, S = \overline{s}).
\end{equation}
Therefore, we report the predictive parity measure as:
\begin{equation}
    PP = |P(Y = +|\hat{Y} = +, S = s) - P(Y = +|\hat{Y} = +, S = \overline{s})|.
\end{equation}
where $PP \in [0, 1]$, with 0 standing for no discrimination and 1 indicating the maximum discrimination. $PP = \dfrac{32}{32+4} - \dfrac{38}{38+5} \approx 0.0052$, in our example.
\subsection{Predictive equality}
\label{subsec:Predictive_equality}
\textit{Predictive equality} \cite{corbett2017algorithmic} (denoted as \emph{PE}), also referred as false positive error (FPR) rate balance \cite{chouldechova2017fair} ($FPR = \dfrac{FP}{TN + FP}$), aims to the equality of decision's accuracy across the protected and non-protected groups. In detail, the probability of students with an actual ``\textit{fail}'' class being incorrectly assigned to the ``\textit{pass}'' class should be the same for both male and female students.
\begin{equation}
P(\hat{Y} = +|Y = -,S = s) = P(\hat{Y} = +|Y = -,S = \overline{s}).    
\end{equation}
In practice, researchers report predictive equality measure by the difference of $FPRs$ \cite{iosifidis2019adafair}:
\begin{equation}
PE = |P(\hat{Y} = +|Y = -,S = s) - P(\hat{Y} = +|Y = -,S = \overline{s})|.
\end{equation}
The value range: $PE \in [0, 1]$, 0 and 1 indicate no discrimination and maximum discrimination, respectively. $PE = |\dfrac{4}{6+4} - \dfrac{5}{5+5}| = 0.1$, in our example.  

\subsection{Treatment equality}
\label{subsec:Treatment_equality}
\emph{Treatment equality} \cite{berk2018fairness} (denoted as \emph{TE}) is satisfied if the ratio of false negatives and false positives is the same for both protected and non-protected groups.
\begin{equation}
\label{eq:treatment_equality}
    \frac{FN_{prot.}}{FP_{prot.}} = \frac{FN_{non-prot.}}{FP_{non-prot.}}.
\end{equation}
In our paper, we report the treatment equality by the difference between two ratios described in Eq.\ref{eq:treatment_equality}.
The metric becomes unbounded if $FP_{prot.}$ or $FP_{non-prot.}$ is zero\footnote{https://docs.aws.amazon.com/sagemaker/latest/dg/clarify-post-training-bias-metric-te.html}. In our example, $TE = -0.2$,  because the ratios of FN and FP are 1 and 1.2 for female and male groups, respectively.

\subsection{Absolute Between-ROC Area}
\label{subsec:ABROCA}
\emph{Absolute Between-ROC Area (ABROCA)}~\cite{gardner2019evaluating} is based on the Receiver Operating Characteristics (ROC) curve. It measures the divergence between the protected ($ROC_{s}$) and non-protected group ($ROC_{\overline{s}}$) curves across all possible thresholds $t \in [0,1]$ of FPR and TPR. The absolute difference between the two curves is measured to capture the case that the curves may cross each other.
\begin{equation}
\label{eq:abroca}
    \int_{0}^{1}\mid ROC_{s}(t) - ROC_{\overline{s}}(t)\mid \,dt.
\end{equation}

The value range: $ABROCA \in [0, 1]$. The lower value indicates a lower difference in the predictions between the two groups and, therefore, a fairer model.
\vspace{-5pt}

\vspace{-5pt}
\section{Evaluation}
\label{sec:evaluation}
\vspace{-5pt}
In this section, we evaluate the performances of predictive models w.r.t. accuracy and fairness measures on five datasets and investigate the effect of choosing grade threshold on fairness measures.
\vspace{-5pt}
\subsection{Datasets}
\label{subsec:datasets}
We evaluate the fairness measures on  popular educational datasets \cite{lequy2022survey,mihaescu2021review,xiao2022survey}, which are summarized in Table \ref{tbl:datasets}. \textcolor{black}{All datasets are imbalanced, as shown in the imbalance ratio (IR) column}.
\vspace{-20pt}
\begin{table}[!htb]
\centering
\caption{An overview of educational datasets}
\label{tbl:datasets}
\begin{adjustbox}{width=1\textwidth}
\begin{tabular}{lrrcccc}
\hline
\multicolumn{1}{c}{\textbf{ Datasets }} &  
\multicolumn{1}{c}{\textbf{ \#Instances}} &
\multicolumn{1}{c}{\begin{tabular}[c]{@{}c@{}}\textbf{ \#Instances}\\\textbf{  (cleaned)} \end{tabular}} &
\multicolumn{1}{c}{\textbf{ \#Attributes }} &
\multicolumn{1}{c}{\begin{tabular}[c]{@{}c@{}}\textbf{ Protected }\\\textbf{ attribute } \end{tabular}} &
\multicolumn{1}{c}{\textbf{Class label}} & 
\textbf{ IR (+:-) } \\ \hline
Law school & 20,798 & 20,798 & 12 & Race & Pass the bar exam & 8.07:1\\
PISA & 5,233 & 3,404 & 24 & Gender & Reading score & 1.35:1 \\
Studden academics & 131 & 131 & 22 & Gender & ESP & 3.70:1\\
Student performance & 649 & 649 & 33 & Gender & Final grade & 5.49:1  \\
xAPI-Edu-Data & 480 & 480 & 17 & Gender & Grade level & 2.78:1 \\
\hline
\end{tabular}
\end{adjustbox}
\vspace{-20pt}
\end{table}

\textbf{Law school}. The Law school dataset\footnote{https://github.com/tailequy/fairness\_dataset/tree/main/Law\_school} contains the law school admission records from 163 law schools in the US in 1991. The target is to predict whether a candidate would pass the bar exam or not. The protected attribute is ``race'' = $\{white,non-white\}$, where \emph{``non-white''} is the protected group.

\textbf{PISA dataset}. The PISA dataset\footnote{https://www.kaggle.com/econdata/pisa-test-scores} contains information on the performance of American students \cite{fleischman2010highlights} taking the exam in 2009 from the Program for International Student Assessment (PISA). The grade threshold (``readingScore'' attribute) is chosen at 500 to compute the class label =  $\{low, high\}$ since the mean reading score is 497.6. The experiments are performed on the cleaned version of this dataset with 3,404 instances after removing missing values.

\textbf{Student academics performance dataset}. The student academics performance dataset\footnote{https://archive.ics.uci.edu/ml/datasets/Student+Academics+Performance} \cite{hussain2018educational} consists of socio-economic, demographic, and academic information of students from three different colleges in India with 22 attributes. The class label is ESP (end semester percentage). In this paper, we encode class label as a binary attribute with values \textit{\{``pass'',``good-and-higher''\}}, where \emph{``good-and-higher''} is a positive class.

\textbf{Student performance dataset}. The student performance dataset\footnote{https://archive.ics.uci.edu/ml/datasets/student+performance} \cite{cortez2008using} was collected in two Portuguese schools in 2005 - 2006. It contains 33 features describing demographics, grades, social and school-related information of students. ``gender'' is considered the protected attribute. The target is to predict the final outcome. The class label = $\{pass, fail\}$ is computed based on the final grade (attribute ``G3'') as \{$<$10, $\geq$10\}\cite{cortez2008using,lequy2022survey}.

\textbf{Students' academic performance dataset (xAPI-Edu-Data)}.
xAPI-Edu-Data\footnote{https://www.kaggle.com/datasets/aljarah/xAPI-Edu-Data} \cite{amrieh2015preprocessing} contains 480 student records described by 17 attributes collected from \textit{Kalboard 360} learning management system. We encode the class label as a binary attribute as $\{Low, Medium-High\}$ corresponding to \textit{\{L, M or H\}} in the original dataset. The positive class is \emph{``Medium-High''}.
\vspace{-7pt}
\subsection{Predictive models}
\label{subsec:models}
\vspace{-2pt}
We select four prevalent classifiers used for student performance prediction problems based on the survey of Xiao et al. \cite{xiao2022survey}, and two well-known fairness-aware classifiers, namely Agarwal's \cite{agarwal2018reductions} and AdaFair \cite{iosifidis2019adafair}. In which, Agarwal's method reduces the fair classification to a sequence of cost-sensitive classification problems with the lowest (empirical) error subject to the desired constraints, and AdaFair is based on AdaBoost that further updates the weights of the instances in each boosting round.
In brief, the predictive models are: 1) Decision Tree (DT); 2) Naive Bayes (NB); 3) Multi-layer Perceptron (MLP); 4) Support Vector Machines (SVM); 5) Agarwal's; 6) AdaFair.
In our experiments, we use 70\% of data for training and 30\% for testing (single split). Predicted models are implemented and executed with default parameters provided by Scikit-learn and Iosifidis et al.~\cite{iosifidis2019adafair}. Agarwal's method is implemented in the AI Fairness 360 toolkit\footnote{https://github.com/Trusted-AI/AIF360}.
\vspace{-7pt}
\subsection{Experimental results}
\label{subsec:results}
\begin{figure*}[!h]
\centering
\vspace{-5pt}
\begin{subfigure}{.32\linewidth}
    \centering
    \includegraphics[width=\linewidth]{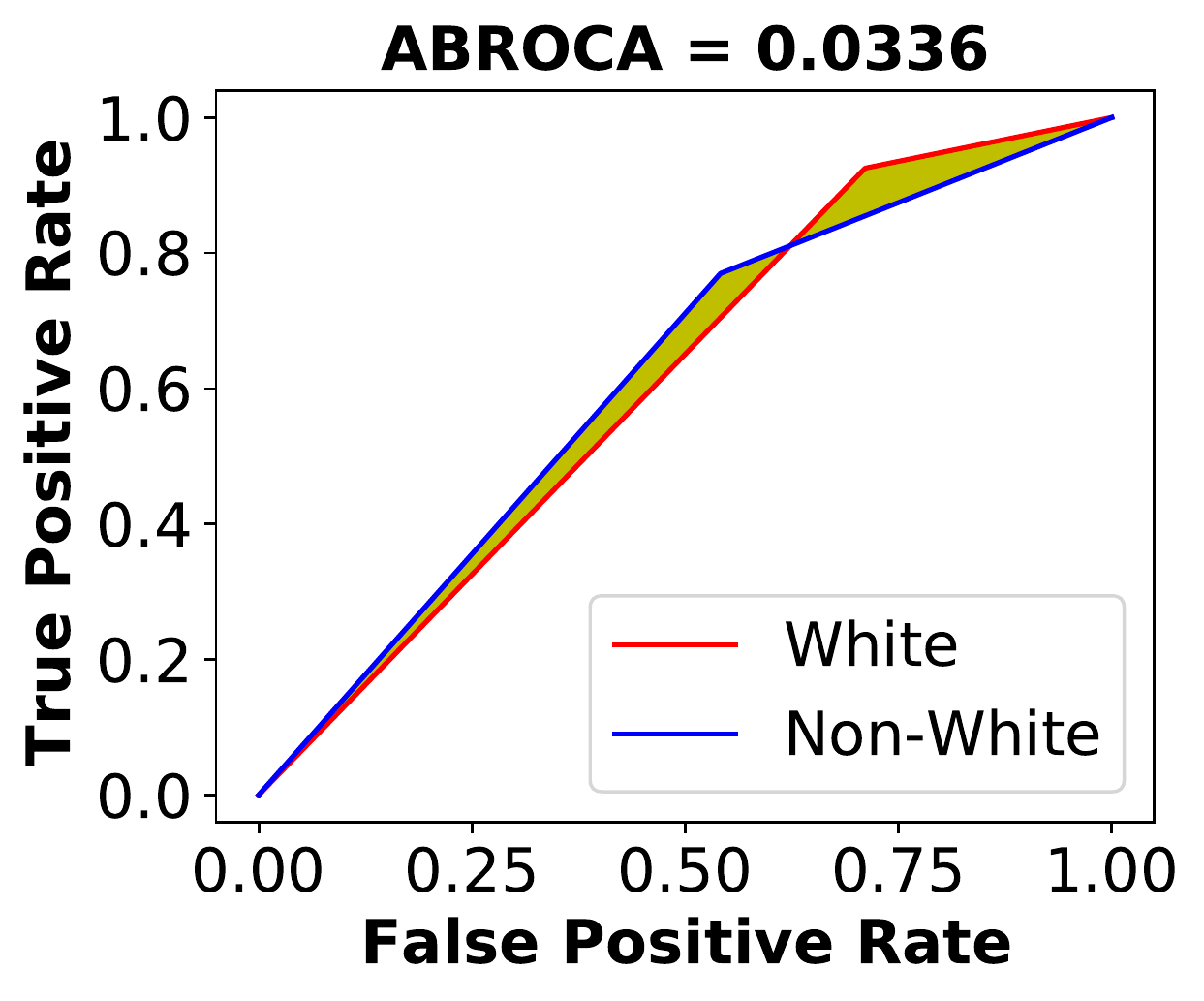}
    \caption{DT}
\end{subfigure}
\hfill
\vspace{-5pt}
\begin{subfigure}{.32\linewidth}
    \centering
    \includegraphics[width=\linewidth]{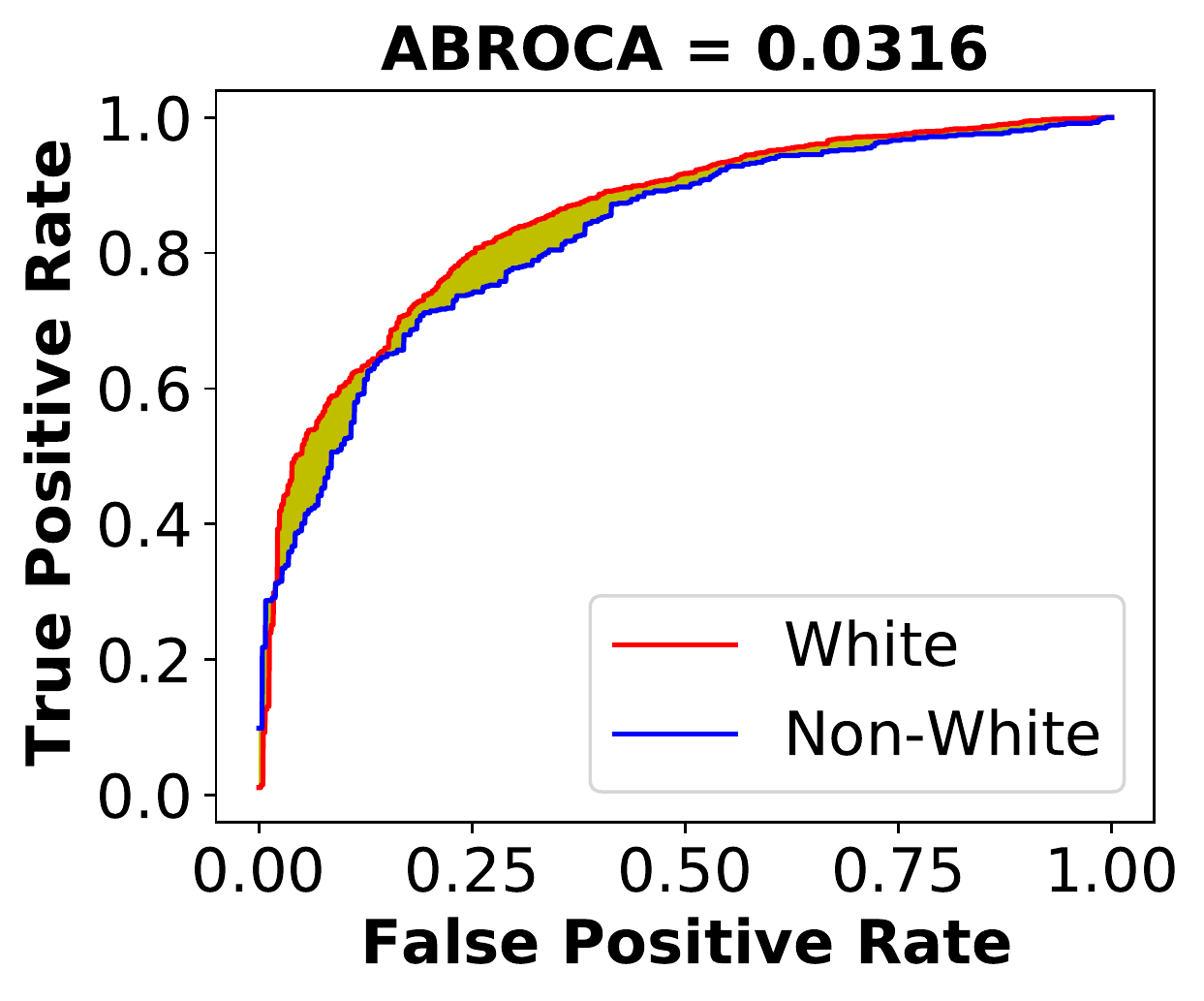}
    \caption{NB}
\end{subfigure}    
\hfill
\begin{subfigure}{.32\linewidth}
    \centering
    \includegraphics[width=\linewidth]{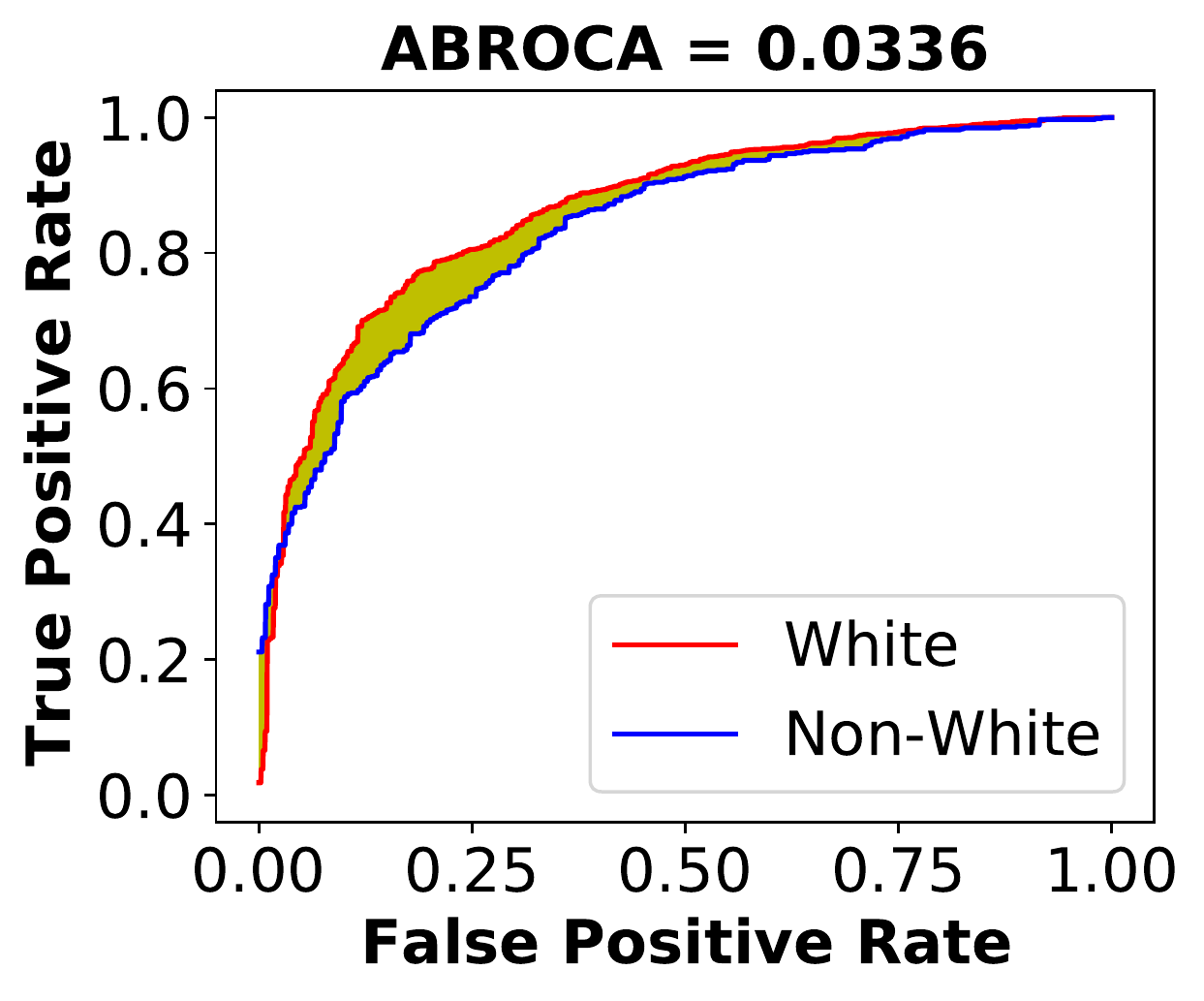}
    \caption{MLP}
\end{subfigure}
\bigskip
\vspace{-5pt}
\begin{subfigure}{.32\linewidth}
    \centering
    \includegraphics[width=\linewidth]{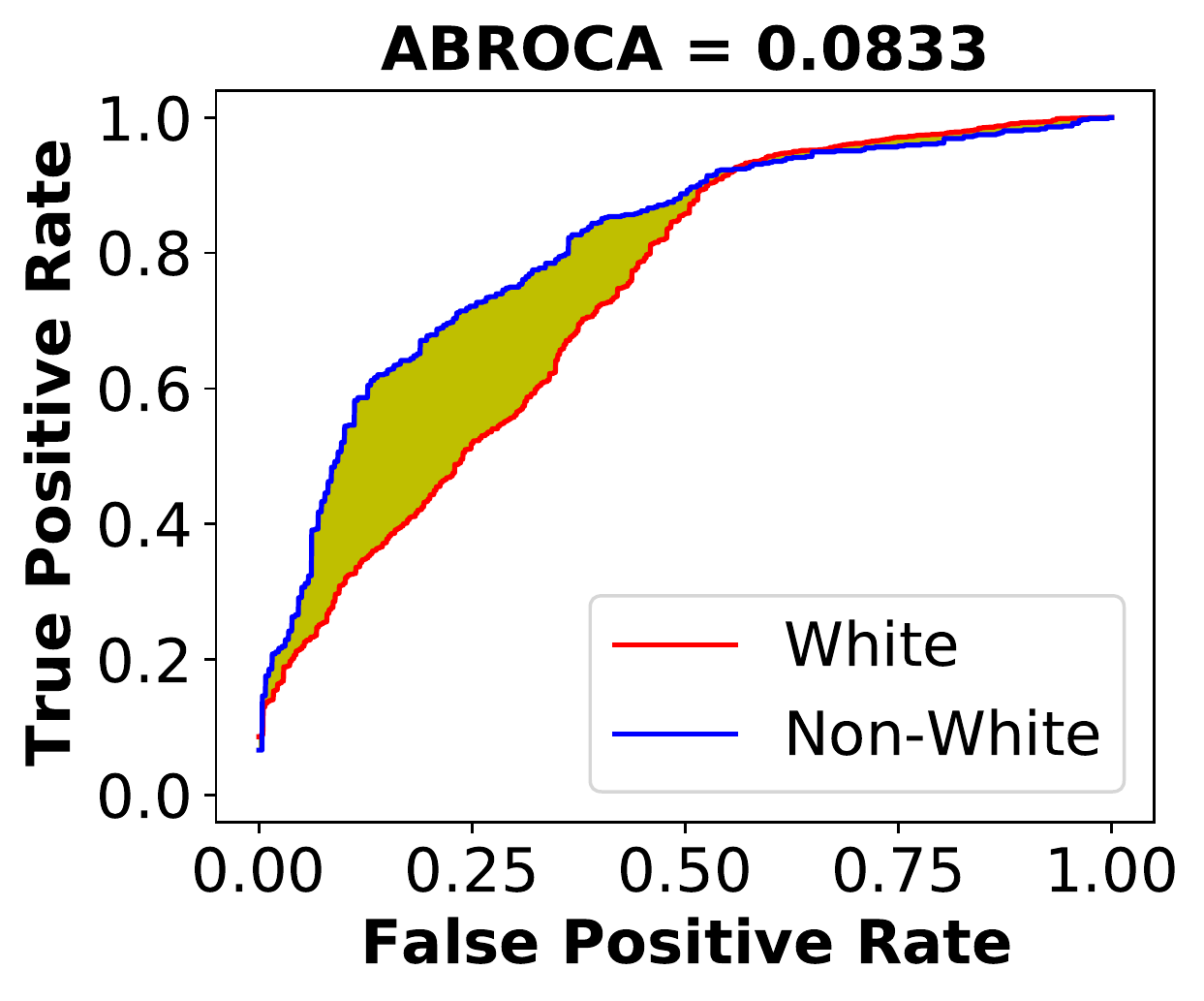}
    \caption{SVM}
\end{subfigure}
\hfill
\vspace{-5pt}
\begin{subfigure}{.32\linewidth}
    \centering
    \includegraphics[width=\linewidth]{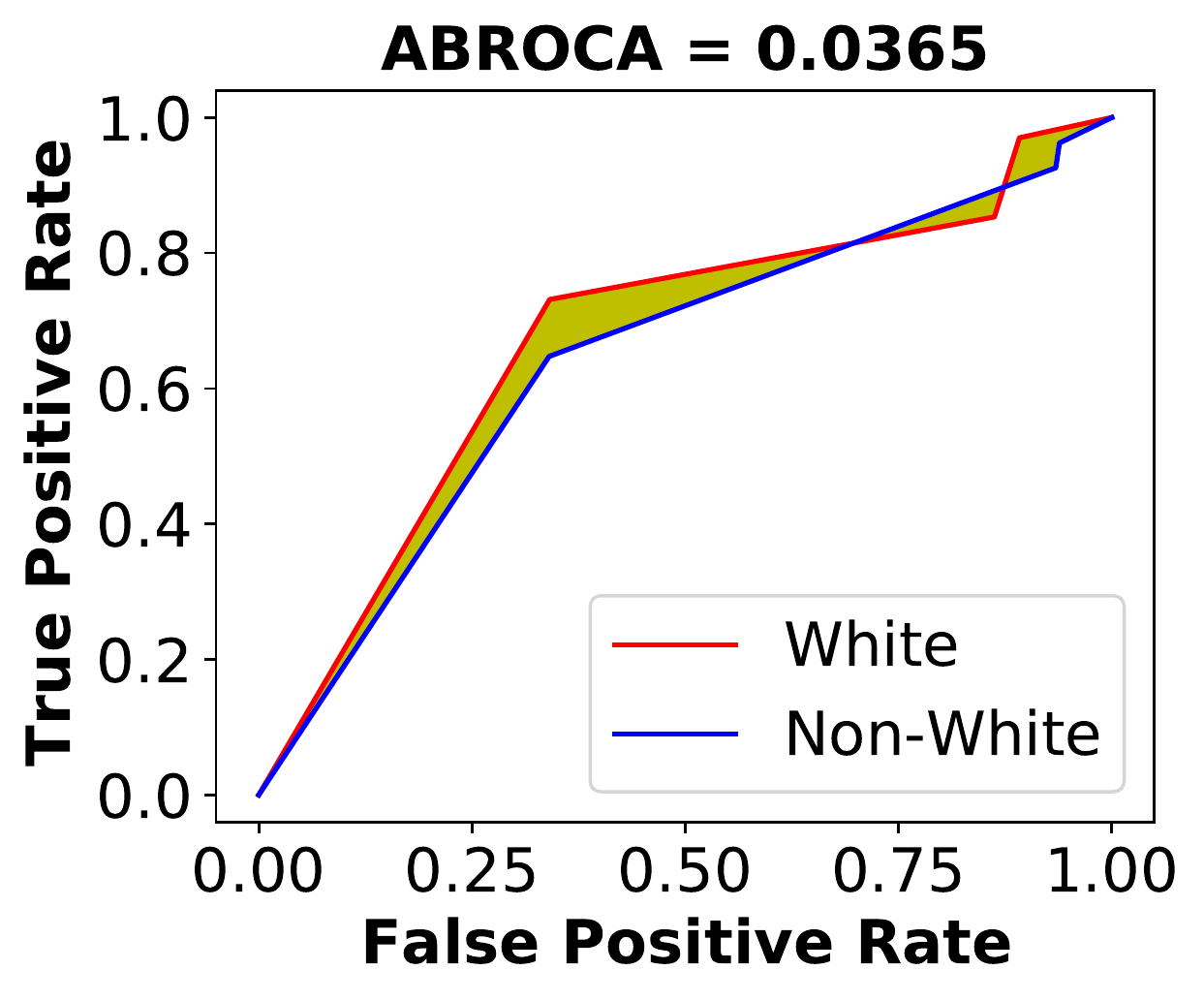}
    \caption{Agarwal's}
\end{subfigure}
\vspace{-5pt}
 \hfill
\begin{subfigure}{.32\linewidth}
    \centering
    \includegraphics[width=\linewidth]{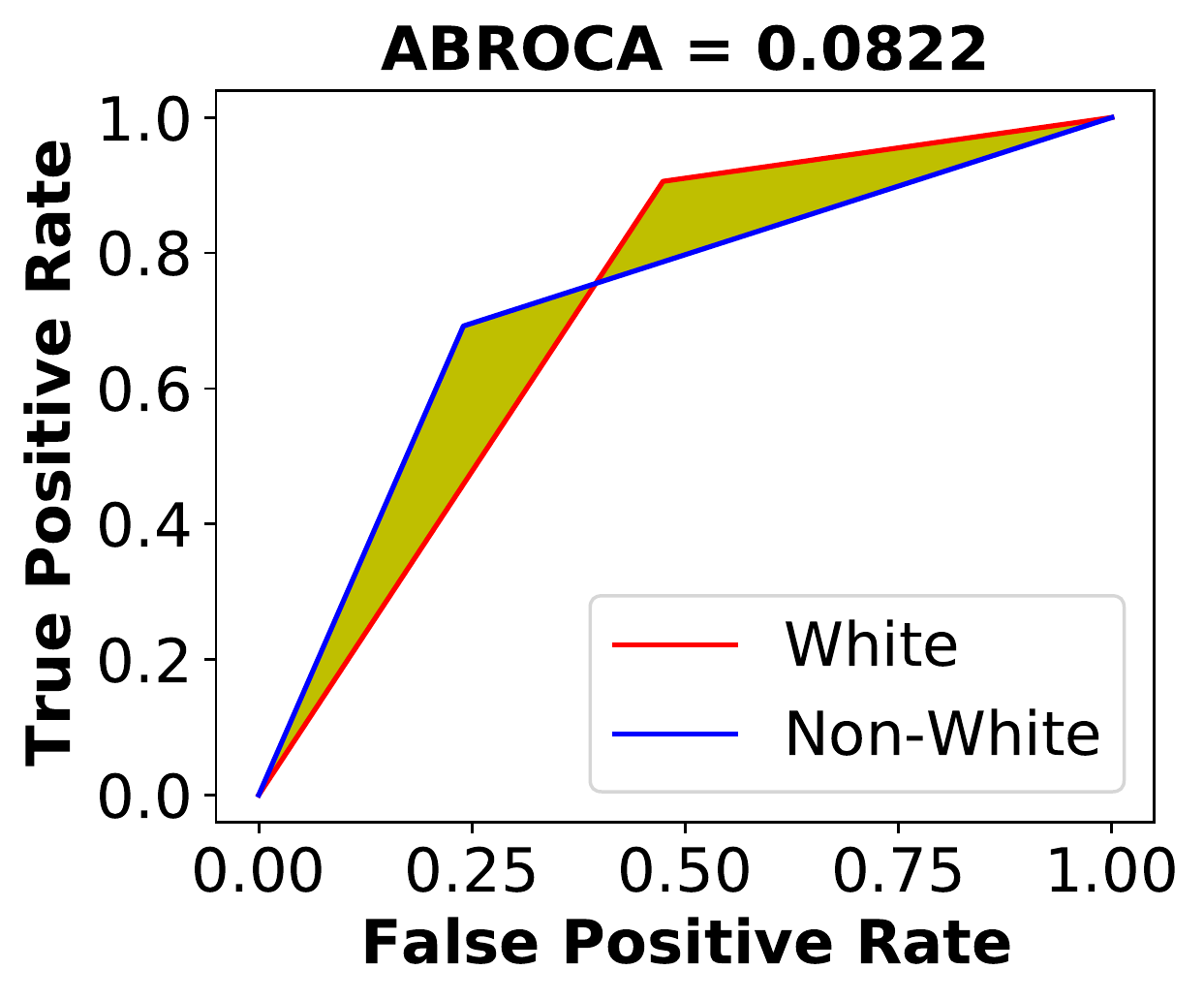}
    \caption{AdaFair}
\end{subfigure}
\caption{Law school: ABROCA slice plots}
\vspace{-20pt}
\label{fig:law_school_abroca}
\end{figure*}

\begin{figure*}[!h]
\centering
\vspace{-5pt}
\begin{subfigure}{.32\linewidth}
    \centering
    \includegraphics[width=\linewidth]{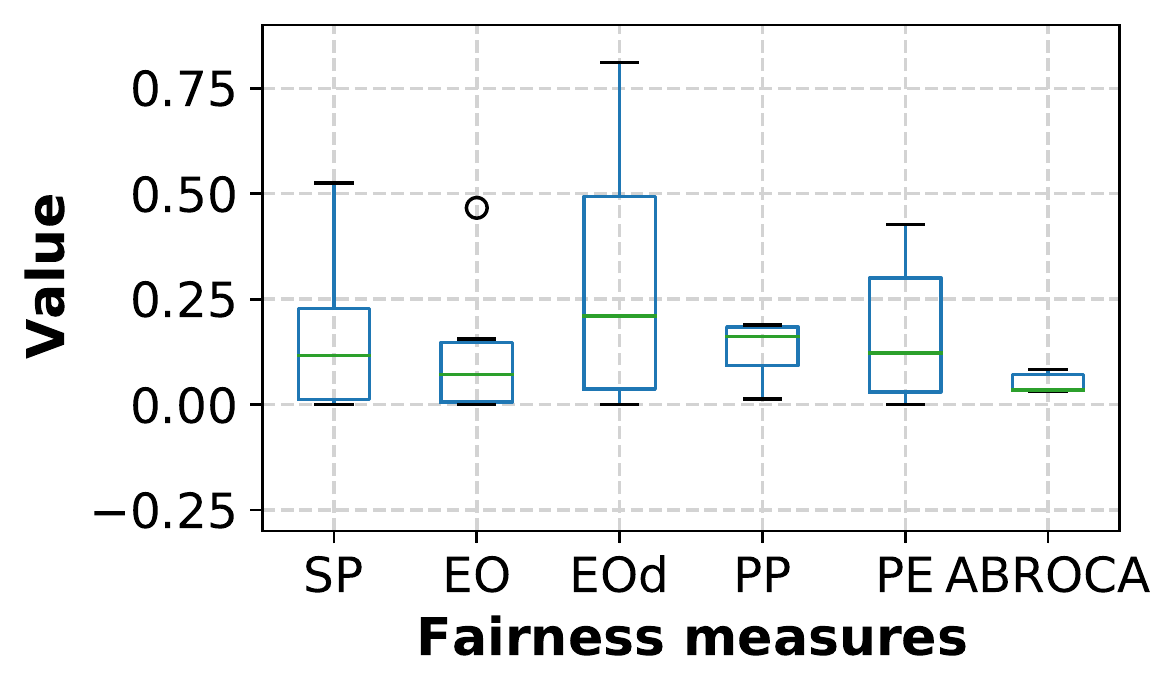}
    \caption{Law school}
\end{subfigure}
\hfill
\vspace{-5pt}
\begin{subfigure}{.32\linewidth}
    \centering
    \includegraphics[width=\linewidth]{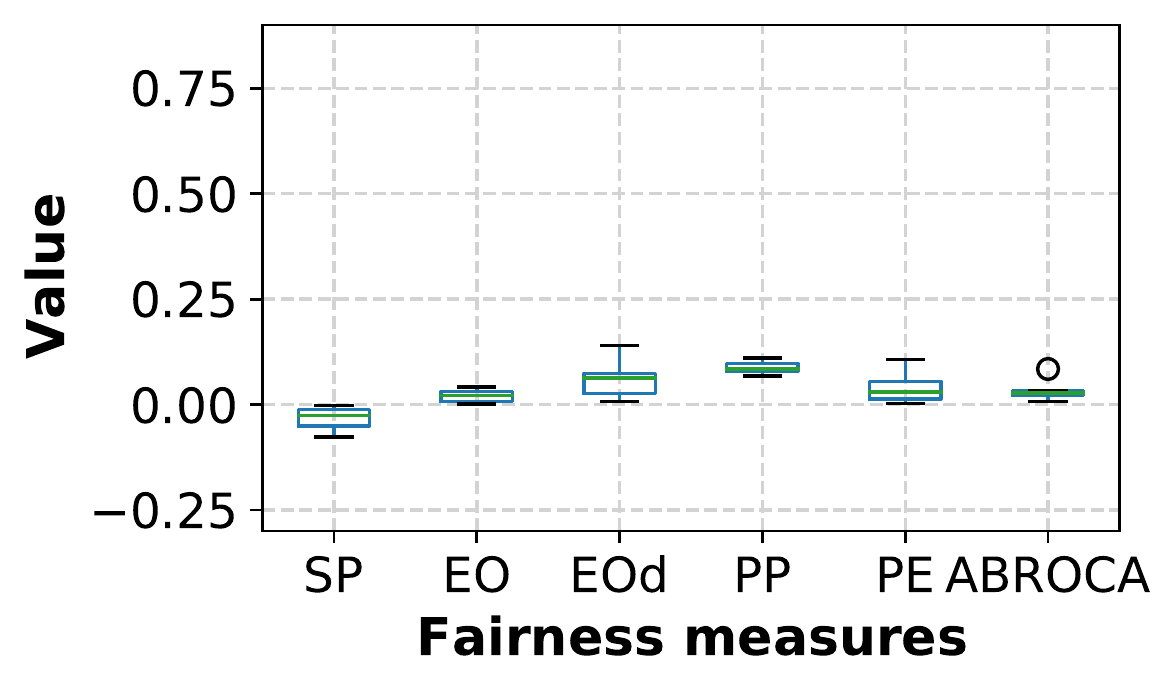}
    \caption{PISA}
\end{subfigure}    
\hfill
\begin{subfigure}{.32\linewidth}
    \centering
    \includegraphics[width=\linewidth]{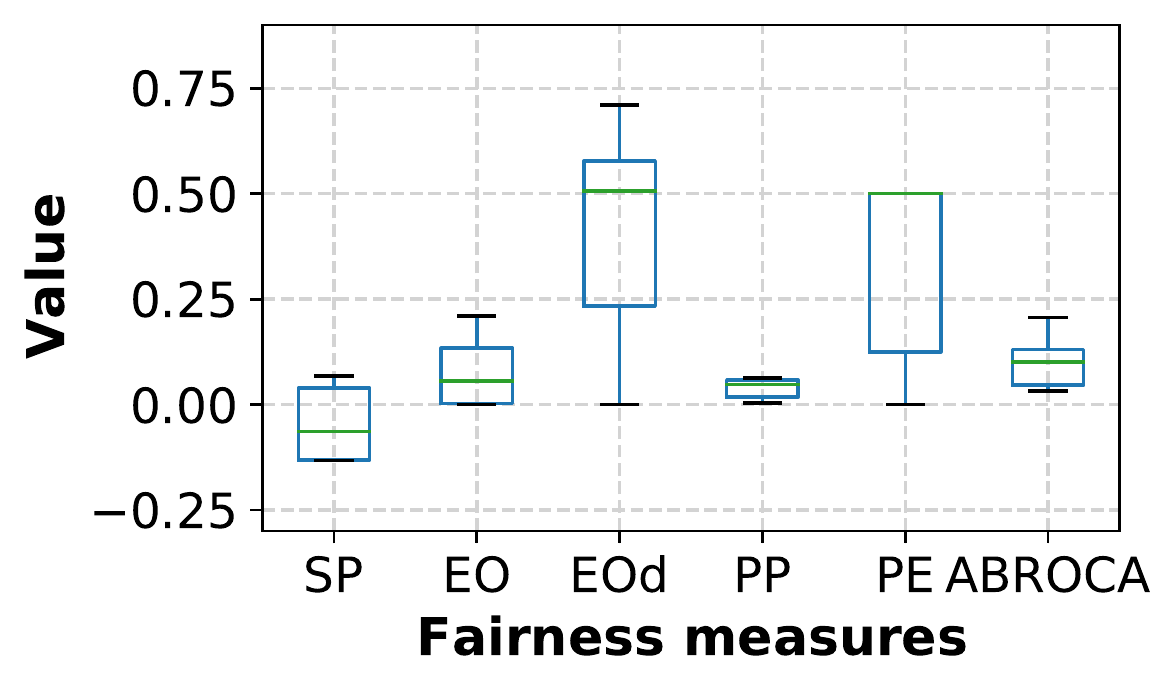}
    \caption{Student academic}
\end{subfigure}
\bigskip
\vspace{-5pt}
\begin{subfigure}{.32\linewidth}
    \centering
    \includegraphics[width=\linewidth]{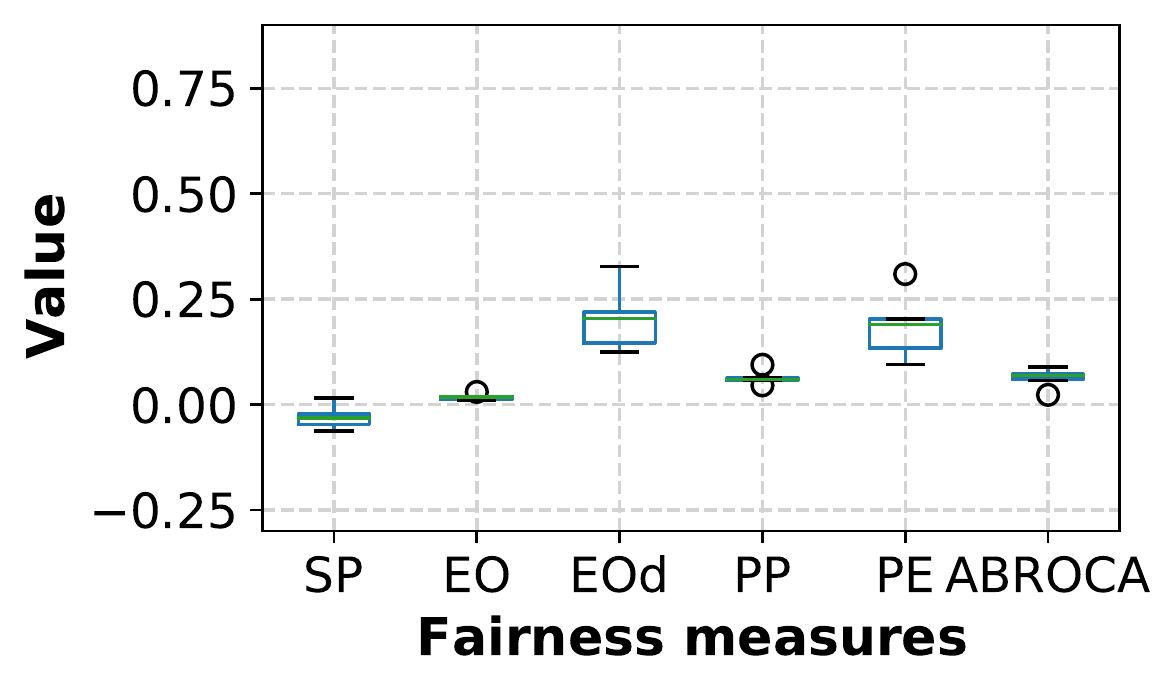}
    \caption{Student performance}
\end{subfigure}
\hfill
\vspace{-5pt}
\begin{subfigure}{.32\linewidth}
    \centering
    \includegraphics[width=\linewidth]{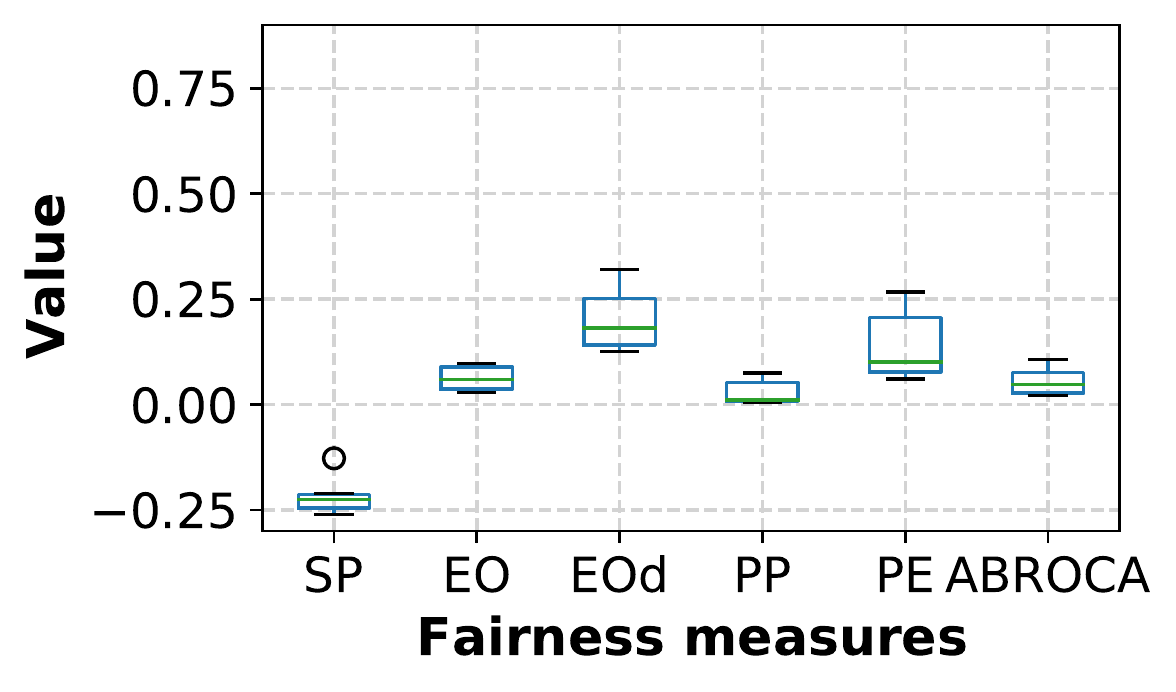}
    \caption{xAPI-Edu-Data}
\end{subfigure}
\vspace{-5pt}
 \hfill
\begin{subfigure}{.32\linewidth}
    \centering
    \includegraphics[width=\linewidth]{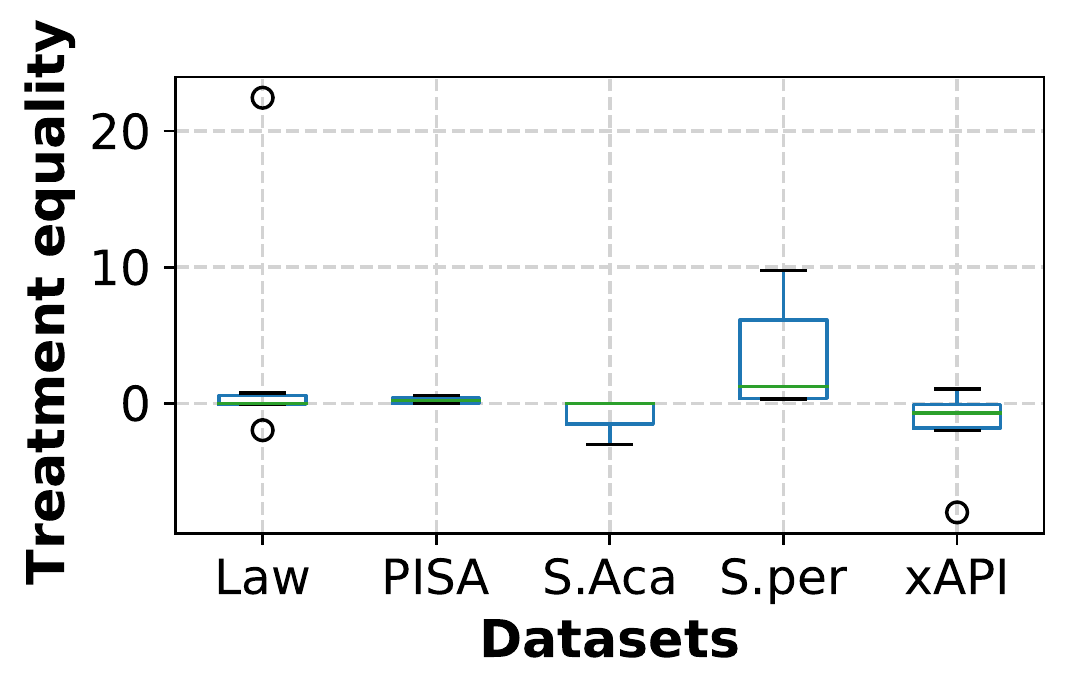}
    \caption{TE measure}
\end{subfigure}
\caption{Variation of fairness measures}
\vspace{-15pt}
\label{fig:Variation_measures}
\end{figure*}

\textbf{Law school dataset.} The results are presented in Table \ref{tbl:law_school}. AdaFair is the best predictive model w.r.t. fairness measures, although its balanced accuracy is significantly lower than that of other models. Besides, the fairness measures show a quite large variation across the classification methods, as demonstrated in Fig. \ref{fig:Variation_measures}-a. Furthermore, the shape and position of the ROC curves, as visualized in Fig. \ref{fig:law_school_abroca}, have been changed across the predictive models, which indicates the change in the performance of models w.r.t. each value in the protected attribute. \textcolor{black}{Because the datasets are imbalanced, we report the performance of predictive models on both accuracy and balanced accuracy measures.}

\begin{table}[!h]
\centering
\caption{Law school: performance of predictive models}\label{tbl:law_school}
\begin{tabular}{lcccccc}
\hline
\textbf{Measures} &  
\multicolumn{1}{c}{\textbf{\phantom{aa}DT\phantom{aa}}} & 
\multicolumn{1}{c}{\textbf{\phantom{aa}NB\phantom{aa}}} &
\multicolumn{1}{c}{\textbf{\phantom{aa}MLP\phantom{aa}}} & 
\multicolumn{1}{c}{\textbf{\phantom{a}SVM}\phantom{a}} &
\multicolumn{1}{c}{\textbf{\phantom{a}Agarwal's\phantom}} &
\multicolumn{1}{c}{\textbf{AdaFair}}\\ \hline
Accuracy            & 0.8458 & 0.8191 & \textbf{0.9042}  & 0.8926 & 0.7952 & 0.8921\\
Balanced accuracy   & 0.6301 & \textbf{0.7784} & 0.6596  & 0.5029 & 0.5848 & 0.5\\
Statistical parity  & 0.1999 & 0.5250 & 0.2367  & 0.0052 & 0.0326 & \textbf{0.0}\\
Equal opportunity   & 0.1557 & 0.4665 & 0.1237  & 0.0014 & 0.0202 & \textbf{0.0}\\
Equalized odds      & 0.3253 & 0.8105 & 0.5501  & 0.0169 & 0.0953 & \textbf{0.0}\\
Predictive parity   & 0.1424 & \textbf{0.0130} & 0.0754  & 0.1857 & 0.1802 & 0.1885\\
Predictive equality & 0.1696 & 0.3440 & 0.4265  & 0.0154 & 0.0751 & \textbf{0.0} \\
Treatment equality  &-0.0667 & 22.440 & 0.7770  & 0.0039 & -1.9676 & \textbf{0.0} \\
ABROCA              & 0.0336 & \textbf{0.0316} & 0.0336  & 0.0833 &  0.0365 & 0.0822\\
\hline
\end{tabular}
\end{table}

\begin{figure*}[!h]
\centering
\vspace{-5pt}
\begin{subfigure}{.32\linewidth}
    \centering
    \includegraphics[width=\linewidth]{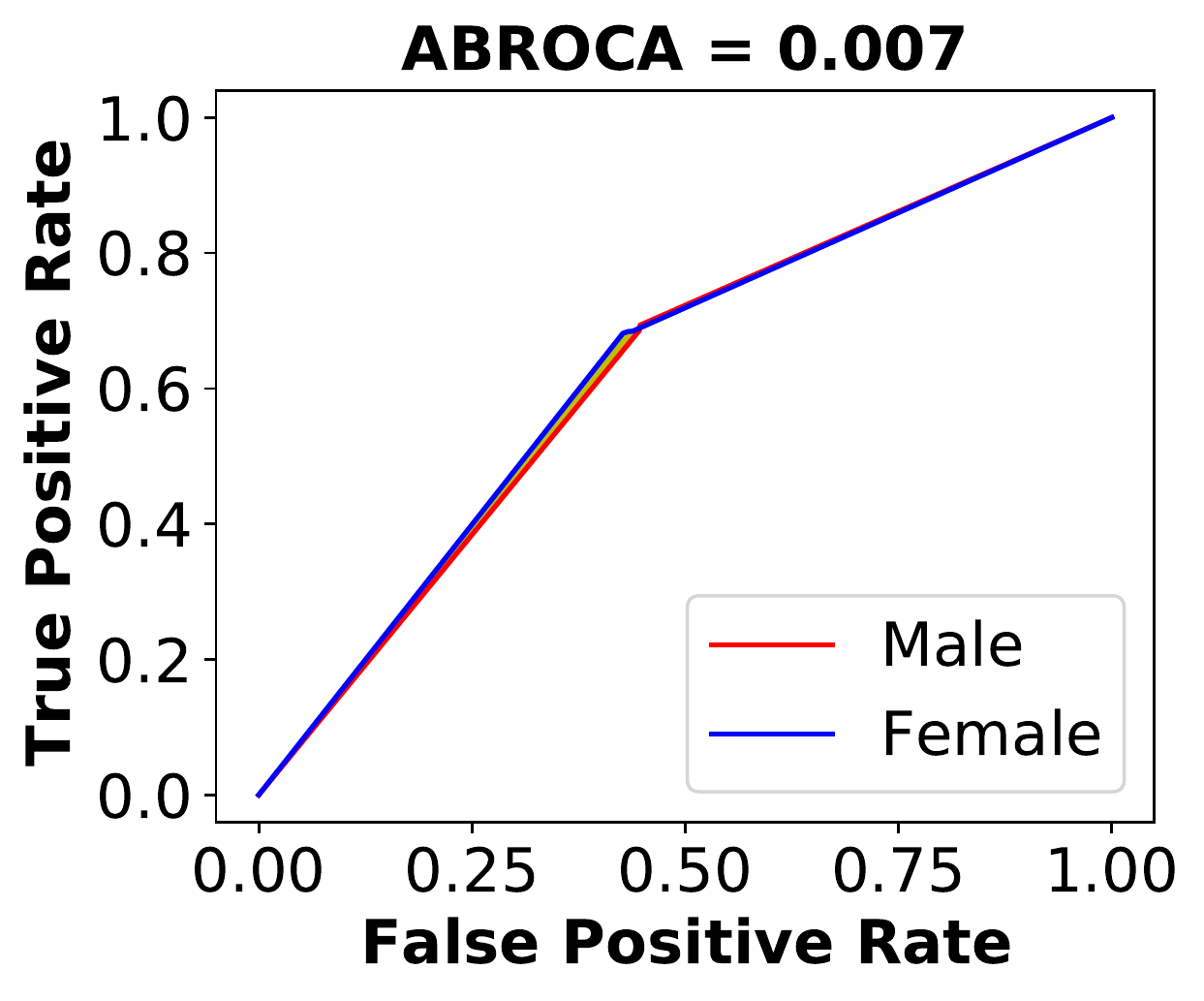}
    \caption{DT}
\end{subfigure}
\hfill
\vspace{-5pt}
\begin{subfigure}{.32\linewidth}
    \centering
    \includegraphics[width=\linewidth]{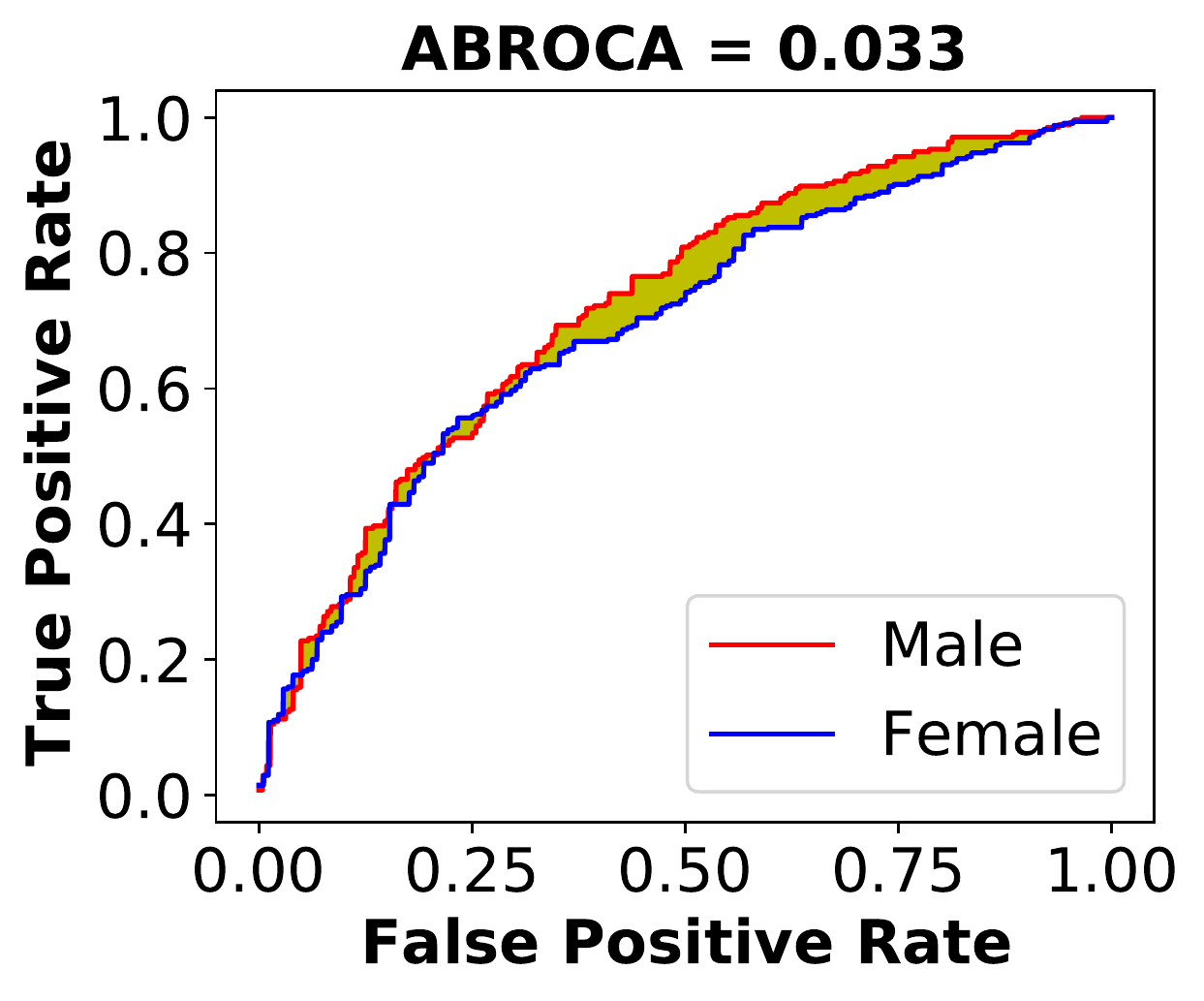}
    \caption{NB}
\end{subfigure}    
\hfill
\begin{subfigure}{.32\linewidth}
    \centering
    \includegraphics[width=\linewidth]{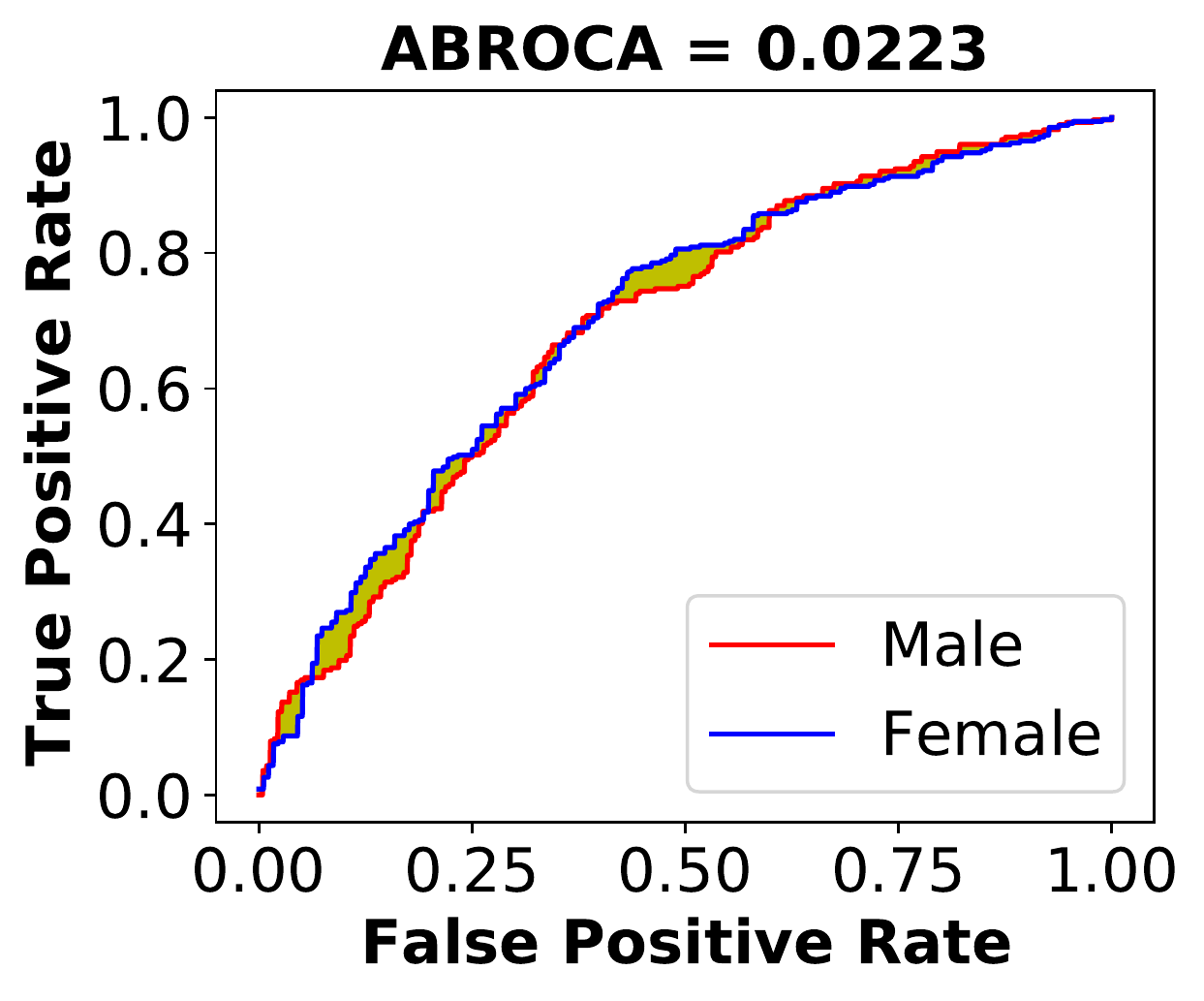}
    \caption{MLP}
\end{subfigure}
\bigskip
\vspace{-5pt}
\begin{subfigure}{.32\linewidth}
    \centering
    \includegraphics[width=\linewidth]{SVM.law.abroca.pdf}
    \caption{SVM}
\end{subfigure}
\hfill
\vspace{-5pt}
\begin{subfigure}{.32\linewidth}
    \centering
    \includegraphics[width=\linewidth]{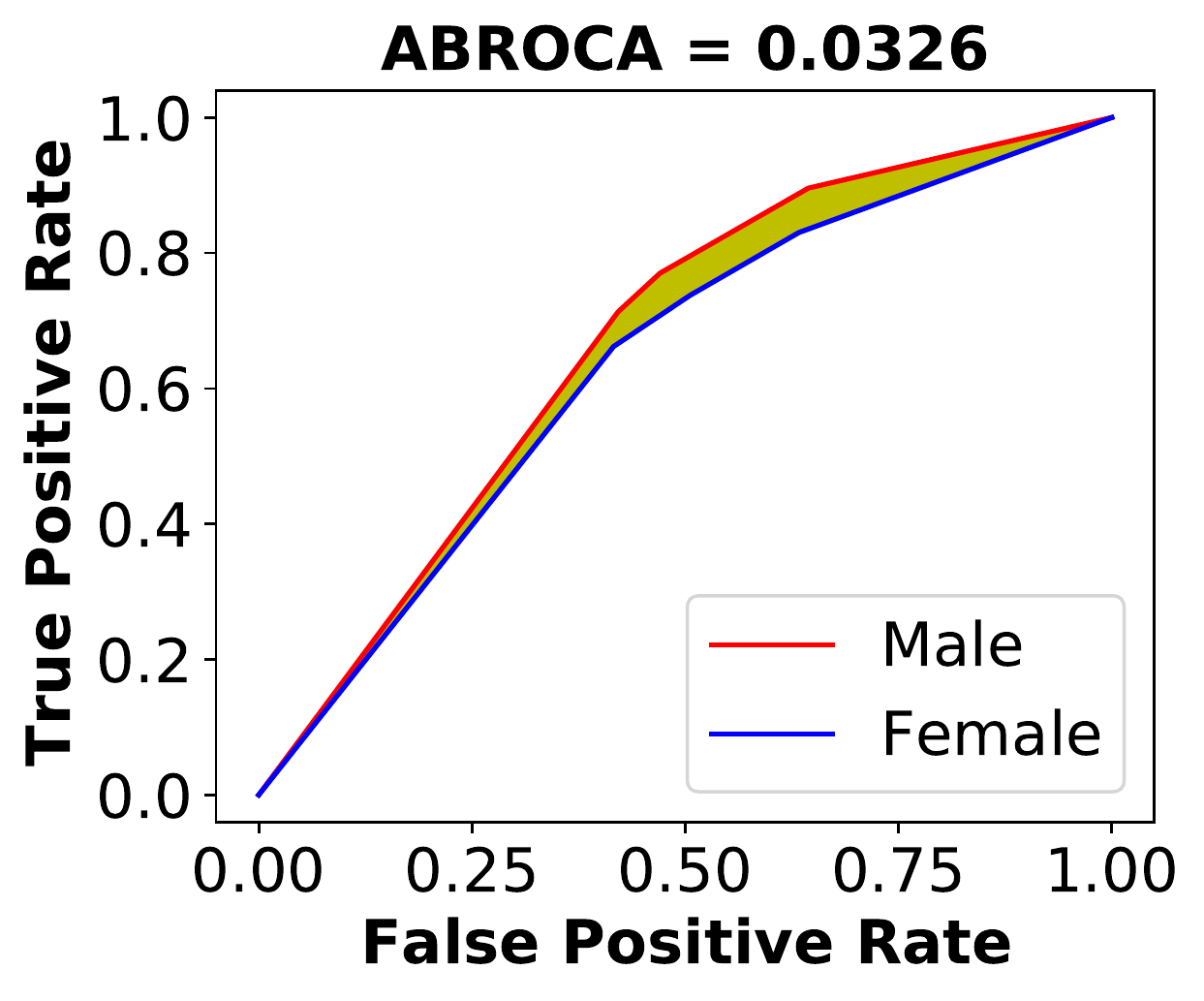}
    \caption{Agarwal's}
\end{subfigure}
\vspace{-5pt}
 \hfill
\begin{subfigure}{.32\linewidth}
    \centering
    \includegraphics[width=\linewidth]{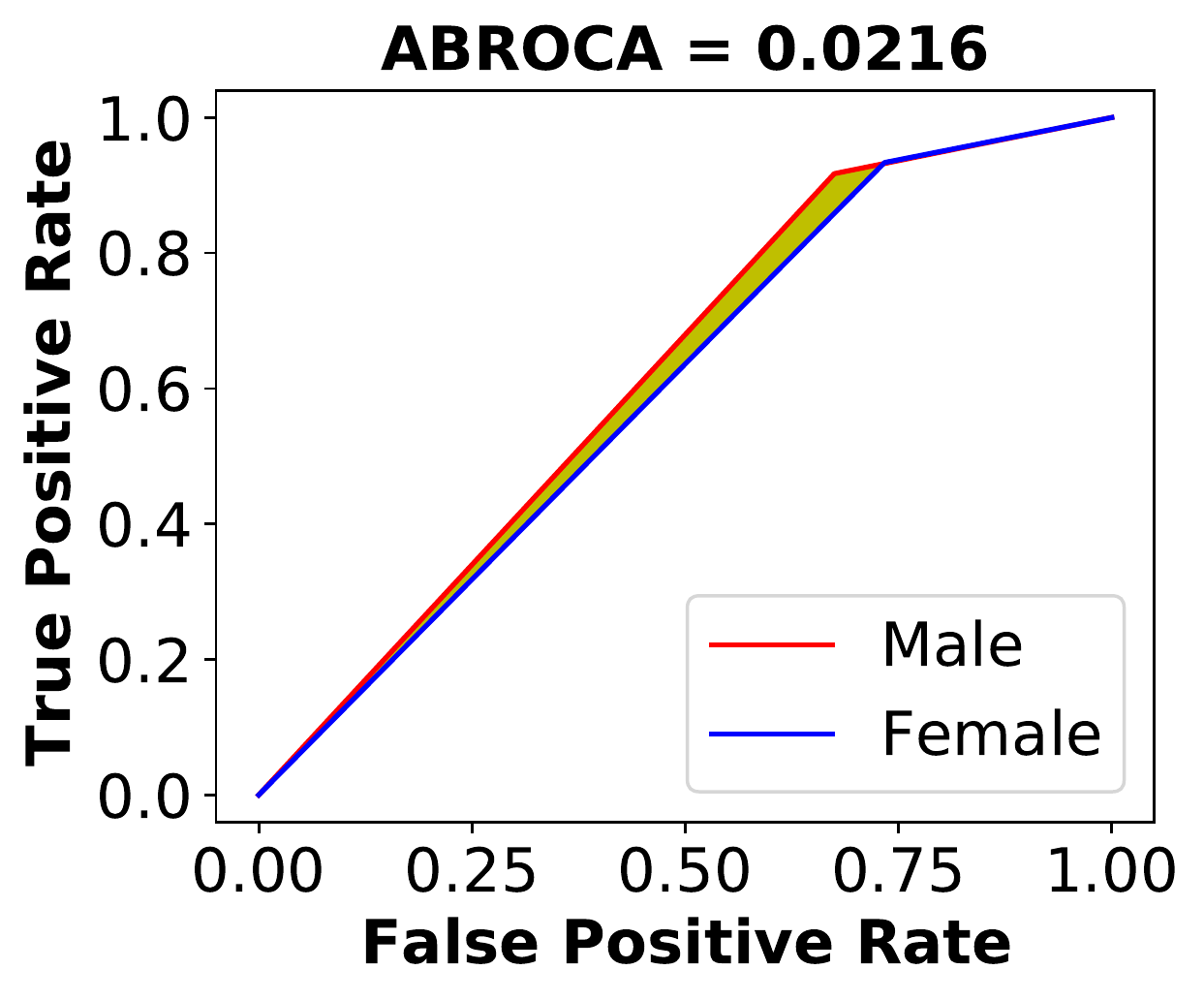}
    \caption{AdaFair}
\end{subfigure}
\caption{PISA: ABROCA slice plots}
\vspace{-10pt}
\label{fig:pisa_abroca}
\end{figure*}

\textbf{PISA dataset.}
\label{subsubsec:pisa}
The interesting point is SVM and DT show their superiority in terms of fairness measures, although AdaFair still has very good results on fairness metrics and accuracy (Fig. \ref{fig:pisa_abroca} and Table \ref{tbl:pisa}). Furthermore, fairness measures have the least variability in this dataset, as shown in Fig. \ref{fig:Variation_measures}-b.
\begin{table}[!h]
\centering
\caption{PISA: performance of predictive models}\label{tbl:pisa}
\begin{tabular}{lcccccc}
\hline
\textbf{Measures} &  
\multicolumn{1}{c}{\textbf{\phantom{aa}DT\phantom{aa}}} & 
\multicolumn{1}{c}{\textbf{\phantom{aa}NB\phantom{aa}}} &
\multicolumn{1}{c}{\textbf{\phantom{aa}MLP\phantom{aa}}} & 
\multicolumn{1}{c}{\textbf{\phantom{a}SVM}\phantom{a}} &
\multicolumn{1}{c}{\textbf{\phantom{a}Agarwal's\phantom}} &
\multicolumn{1}{c}{\textbf{AdaFair}}\\ \hline
Accuracy            & 0.6360  & 0.6624  & 0.6526 & 0.6096 & 0.6614 & \textbf{0.6810}\\
Balanced accuracy   & 0.6224  & \textbf{0.6379}  & 0.5732    & 0.5026 & 0.6340 & 0.6130\\
Statistical parity  & -0.0200 & -0.0316 & -0.0771  & \textbf{-0.0022}& -0.0096 & -0.0573\\
Equal opportunity   & \textbf{0.0019}  & 0.0262  & 0.0330    & 0.0043 & 0.0414 & 0.0164\\
Equalized odds      & 0.0165  & 0.0709  & 0.1398    & \textbf{0.0068} & 0.0548 & 0.0752\\
Predictive parity   & 0.1012  & \textbf{0.0683}  & 0.0826    & 0.1108 & 0.0785 & 0.0868\\
Predictive equality & 0.0146  & 0.0446  & 0.1067   & \textbf{0.0024} & 0.0134 & 0.0588\\
Treatment equality  & 0.5642  & 0.3855  & -0.0251   & \textbf{-0.0033}& 0.4609 & 0.0260\\
ABROCA              & \textbf{0.0070}  & 0.0330  & 0.0223    & 0.0844 & 0.0326 & 0.0216\\
\hline
\end{tabular}
\vspace{-10pt}
\end{table}

\begin{figure*}[!h]
\centering
\vspace{-5pt}
\begin{subfigure}{.32\linewidth}
    \centering
    \includegraphics[width=\linewidth]{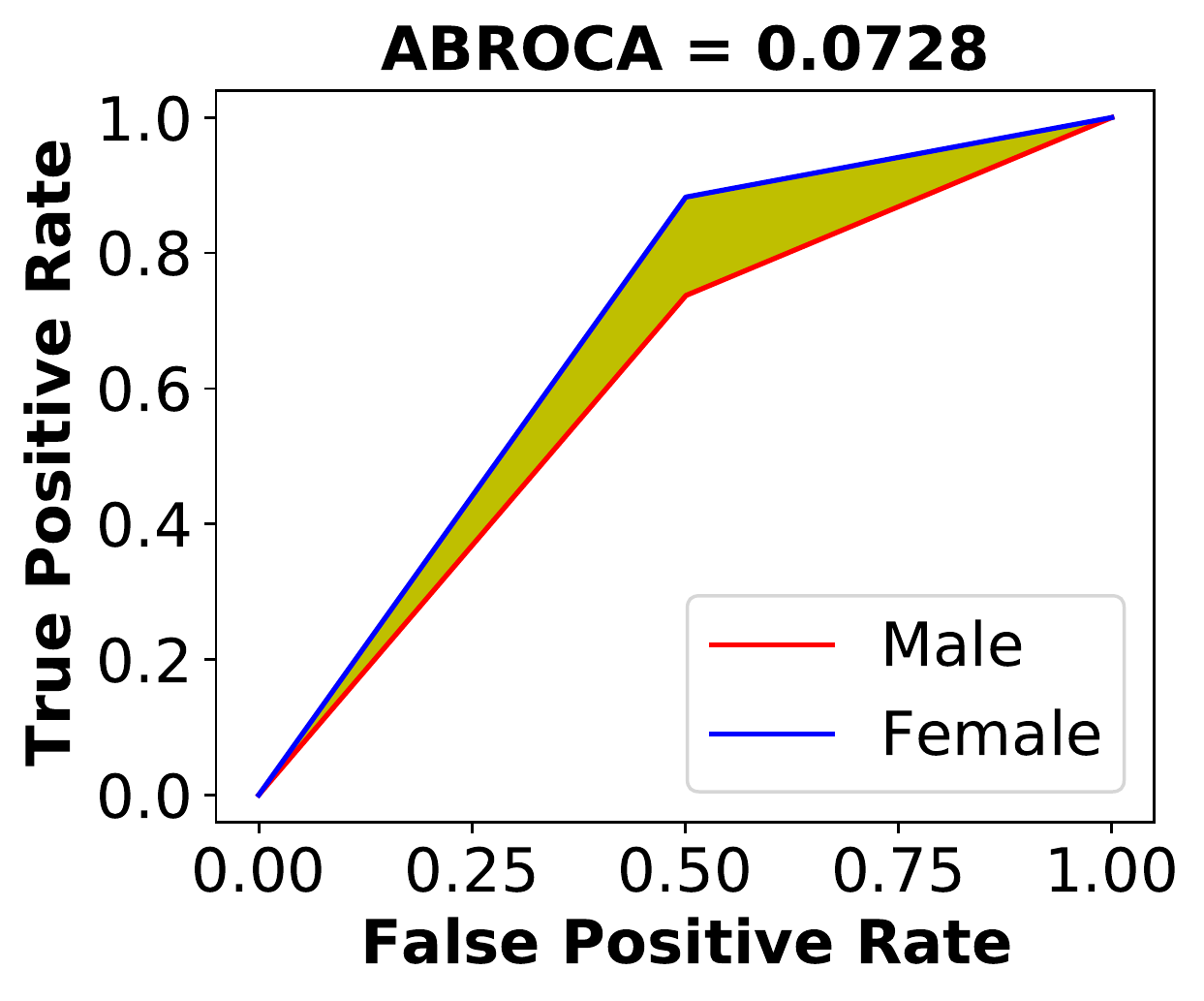}
    \caption{DT}
\end{subfigure}
\hfill
\vspace{-5pt}
\begin{subfigure}{.32\linewidth}
    \centering
    \includegraphics[width=\linewidth]{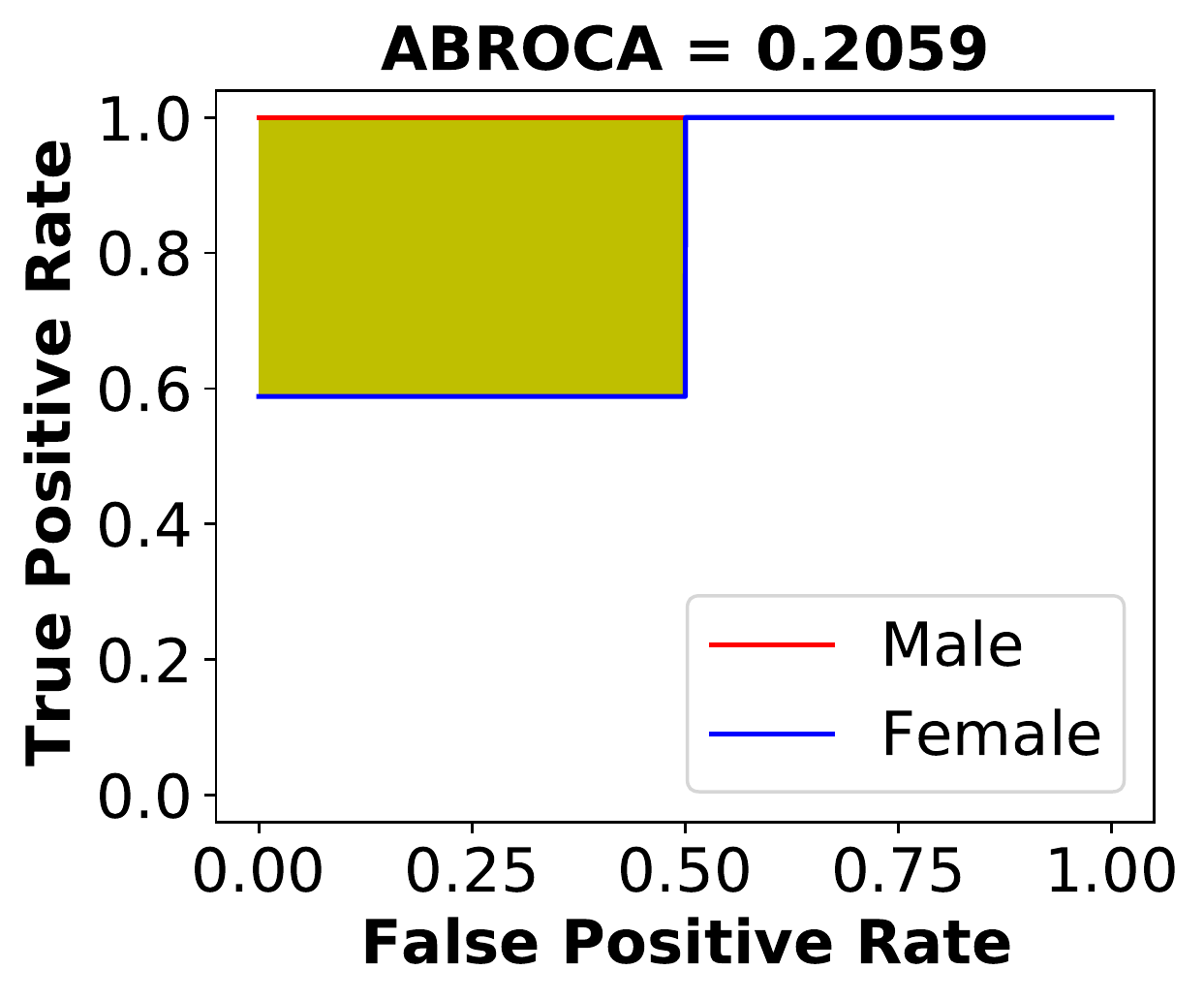}
    \caption{NB}
\end{subfigure}    
\hfill
\begin{subfigure}{.32\linewidth}
    \centering
    \includegraphics[width=\linewidth]{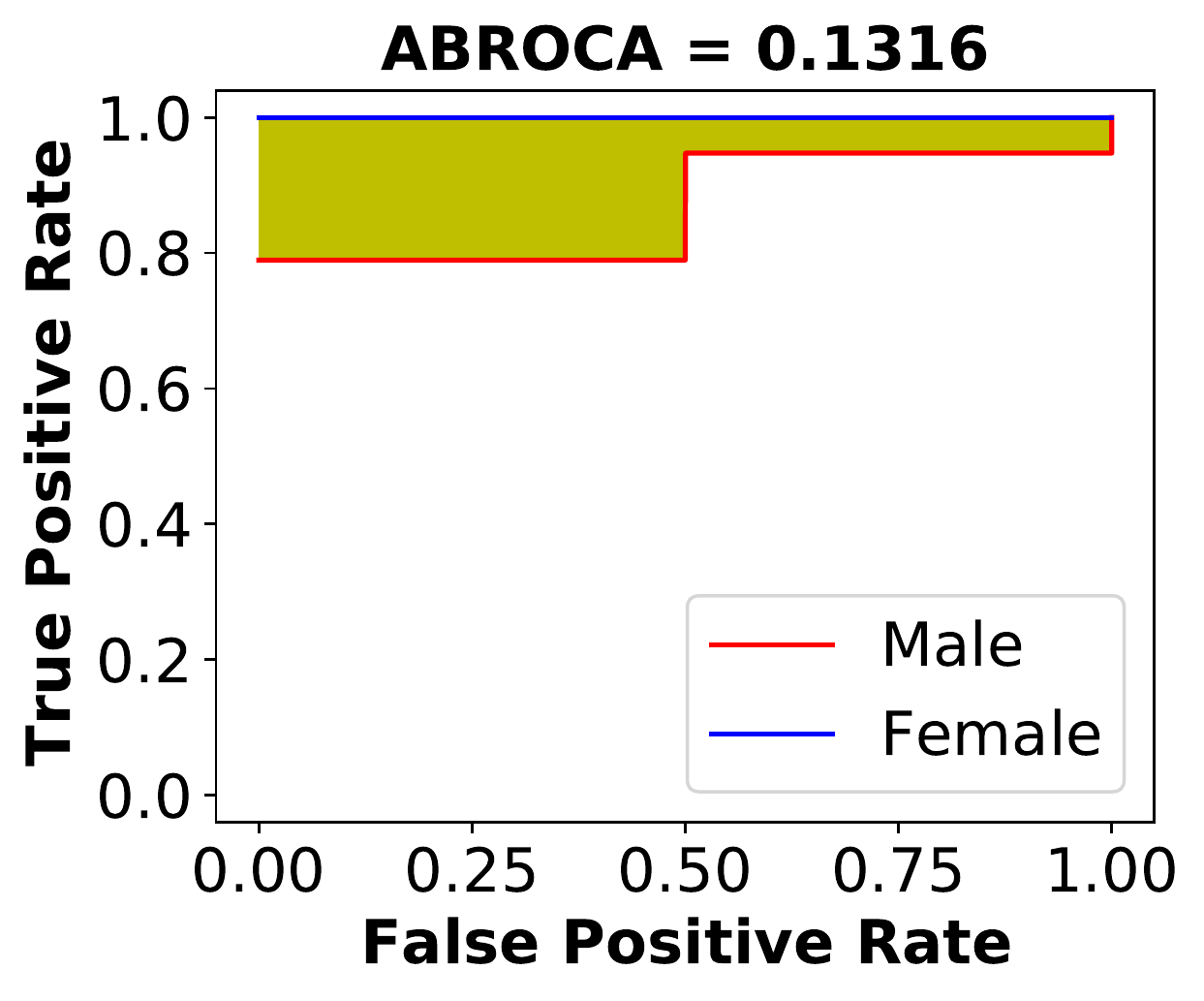}
    \caption{MLP}
\end{subfigure}
\bigskip
\vspace{-5pt}
\begin{subfigure}{.32\linewidth}
    \centering
    \includegraphics[width=\linewidth]{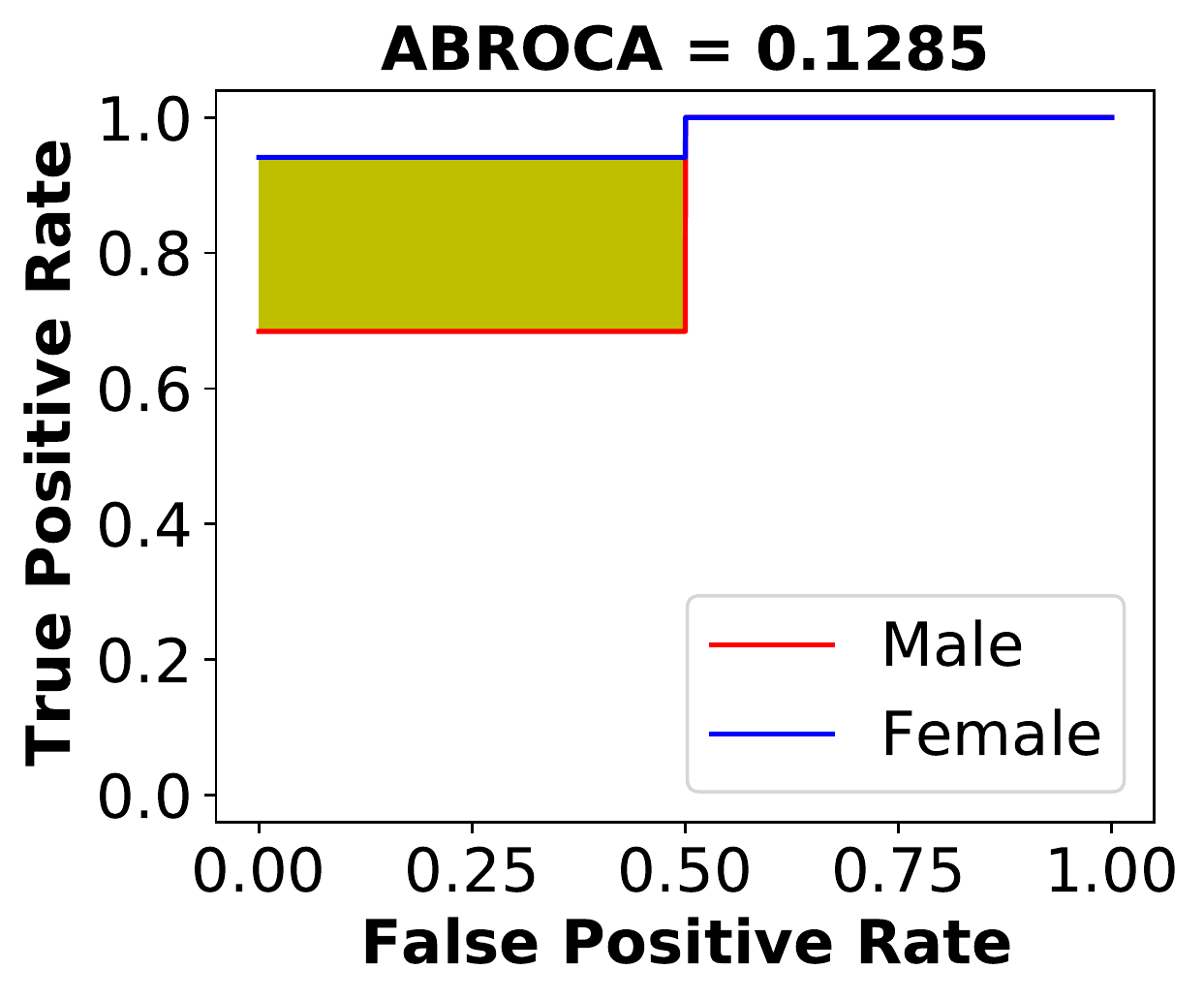}
    \caption{SVM}
\end{subfigure}
\hfill
\vspace{-5pt}
\begin{subfigure}{.32\linewidth}
    \centering
    \includegraphics[width=\linewidth]{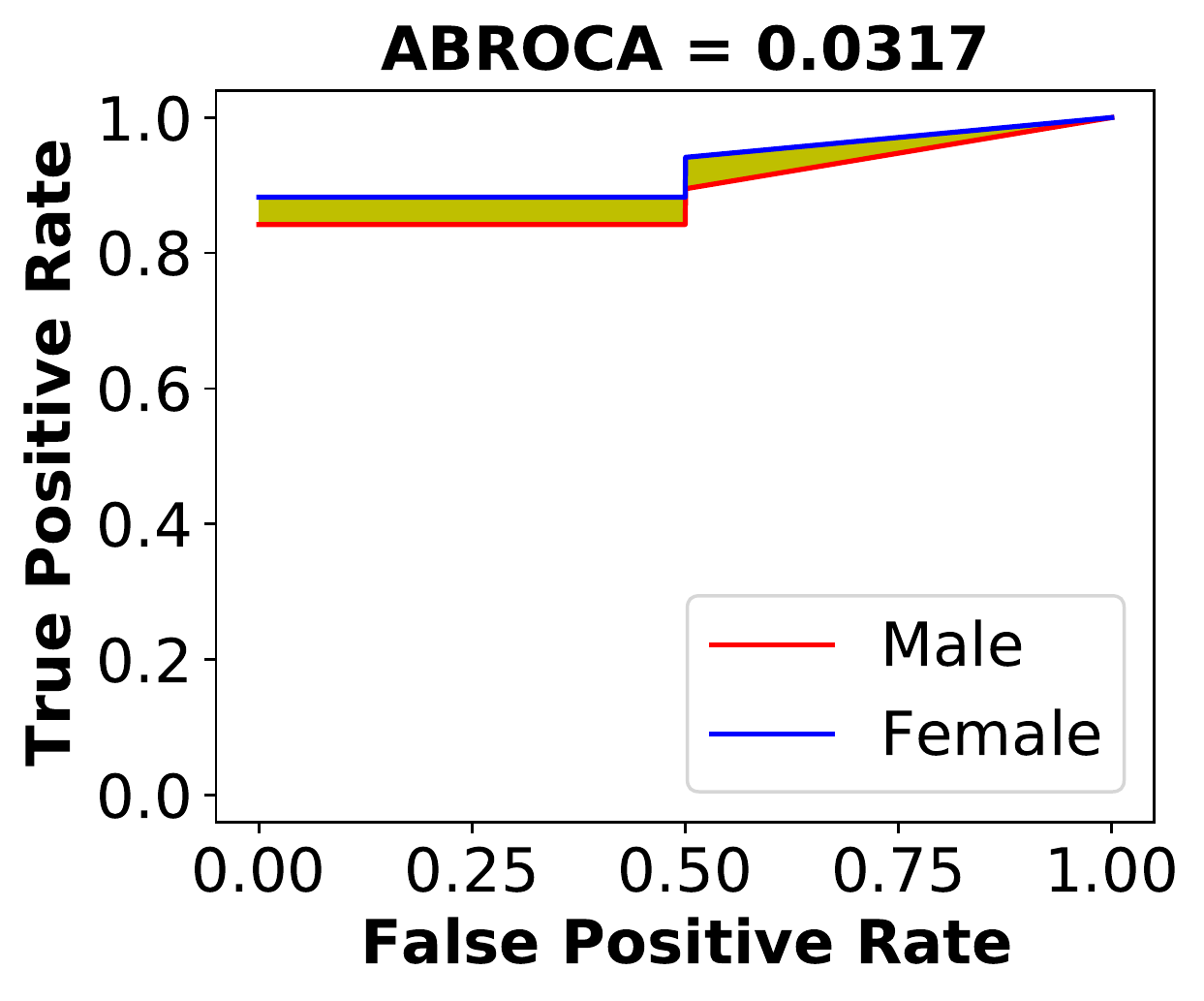}
    \caption{Agarwal's}
\end{subfigure}
\vspace{-5pt}
 \hfill
\begin{subfigure}{.32\linewidth}
    \centering
    \includegraphics[width=\linewidth]{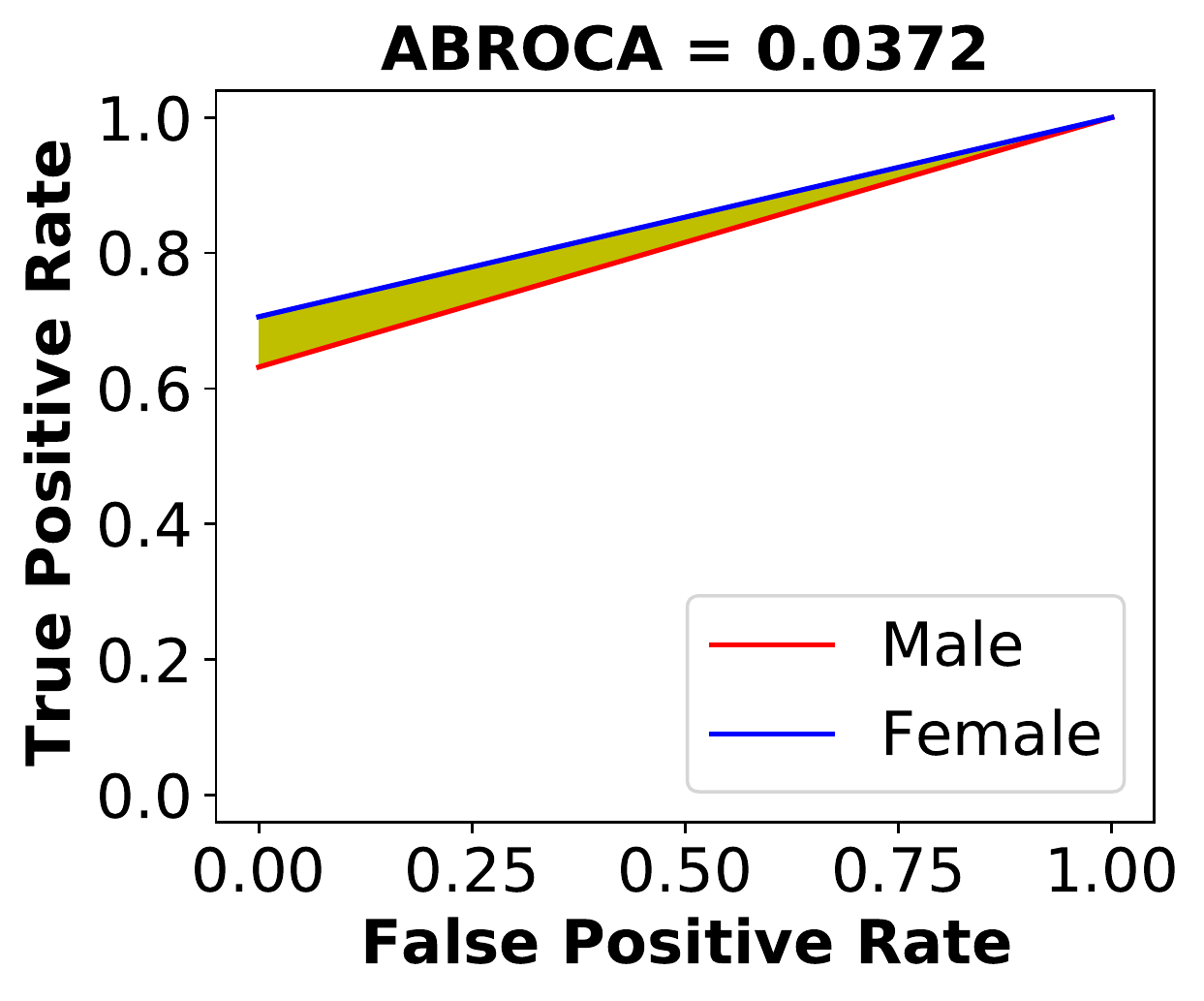}
    \caption{AdaFair}
\end{subfigure}
\caption{Student academics: ABROCA slice plots}
\vspace{-10pt}
\label{fig:student_academics_abroca}
\end{figure*}

\textbf{Student academics performance dataset}.
\label{subsubsec:student_academics}
The AdaFair outperforms other models w.r.t. fairness measures, however, the balanced accuracy is decreased considerably (Table \ref{tbl:student_academic}). Besides, all fairness measures have significant variation across predictive models (Fig. \ref{fig:student_academics_abroca} and \ref{fig:Variation_measures}-c).
\begin{table}[!hb]
\centering
\vspace{-20pt}
\caption{Student academics: performance of predictive models}\label{tbl:student_academic}
\begin{tabular}{lcccccc}
\hline
\textbf{Measures} &  
\multicolumn{1}{c}{\textbf{\phantom{aa}DT\phantom{aa}}} & 
\multicolumn{1}{c}{\textbf{\phantom{aa}NB\phantom{aa}}} &
\multicolumn{1}{c}{\textbf{\phantom{aa}MLP\phantom{aa}}} & 
\multicolumn{1}{c}{\textbf{\phantom{a}SVM}\phantom{a}} &
\multicolumn{1}{c}{\textbf{\phantom{a}Agarwal's\phantom}} &
\multicolumn{1}{c}{\textbf{AdaFair}}\\ \hline
Accuracy            & 0.7750  & 0.8750  & 0.8750 &  \textbf{0.9250} & 0.8750 & 0.9\\
Balanced accuracy   & 0.6528  & \textbf{0.8194}  & \textbf{0.8194}  & 0.6250 & \textbf{0.8194} & 0.5\\
Statistical parity  & -0.1278 &-0.1328  &-0.1328  & 0.0526 & 0.0677 & \textbf{0.0}\\
Equal opportunity   & 0.1455  & 0.0991  & 0.2105    & \textbf{0.0}    & 0.0123 & \textbf{0.0}\\
Equalized odds      & 0.1455  & 0.5991  & 0.7105     & 0.5    & 0.5124 & \textbf{0.0}\\
Predictive parity   & \textbf{0.0042}  & 0.0588   & 0.0552 & 0.0397 & 0.0556 & 0.01\\
Predictive equality & \textbf{0.0}     & 0.5     & 0.5        & 0.5    & 0.5 & \textbf{0.0}\\
Treatment equality  & -3.0    &$N/A$    & $N/A$      & \textbf{0.0}     &$N/A$ & \textbf{0.0}\\
ABROCA              & 0.0728  & 0.2059  & 0.1316 &  0.1285 & \textbf{0.0317} & 0.0372\\
\hline
\end{tabular}
\vspace{-10pt}
\end{table}

\begin{figure*}[!h]
\centering
\vspace{-5pt}
\begin{subfigure}{.32\linewidth}
    \centering
    \includegraphics[width=\linewidth]{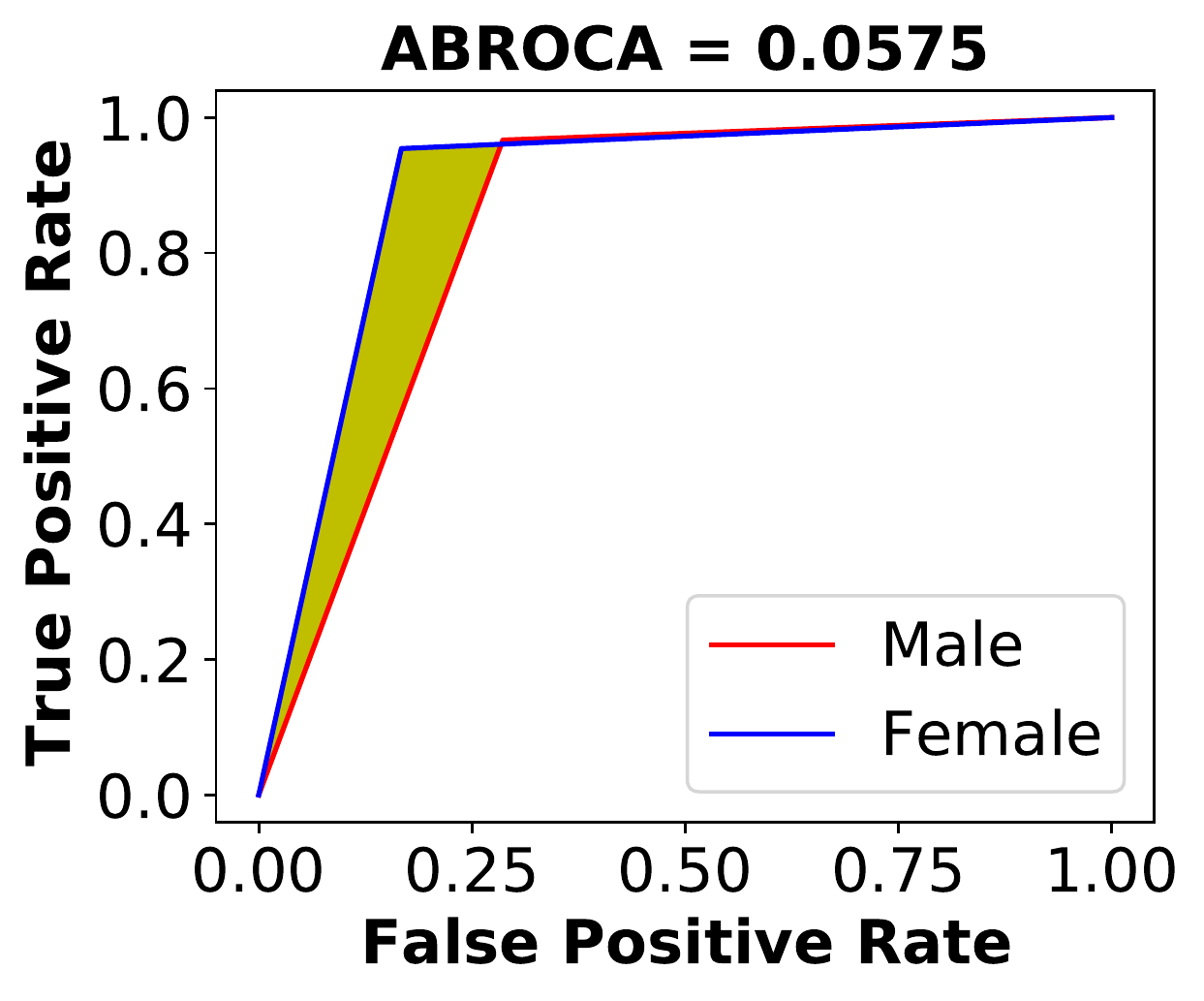}
    \caption{DT}
\end{subfigure}
\hfill
\vspace{-5pt}
\begin{subfigure}{.32\linewidth}
    \centering
    \includegraphics[width=\linewidth]{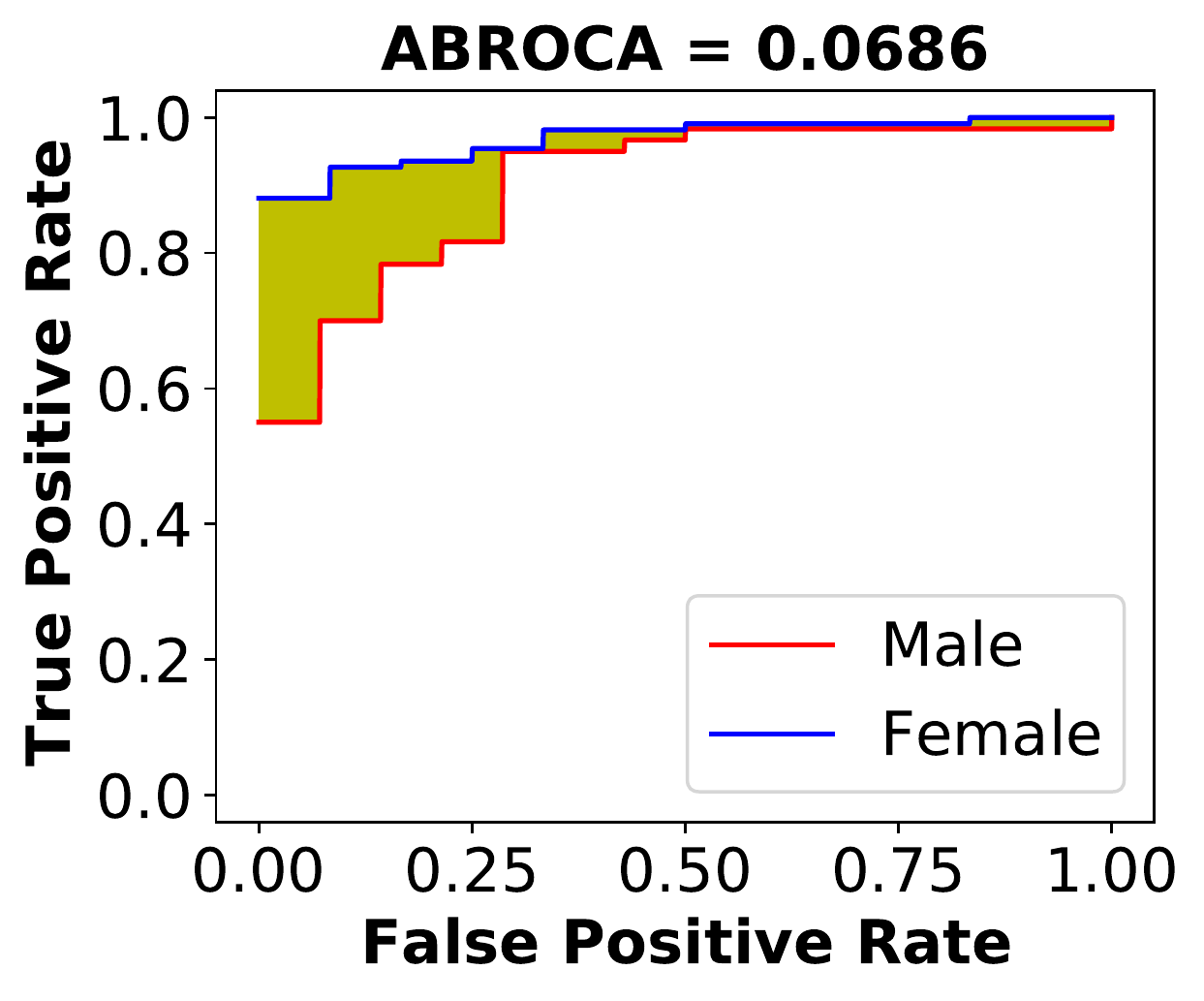}
    \caption{NB}
\end{subfigure}    
\hfill
\begin{subfigure}{.32\linewidth}
    \centering
    \includegraphics[width=\linewidth]{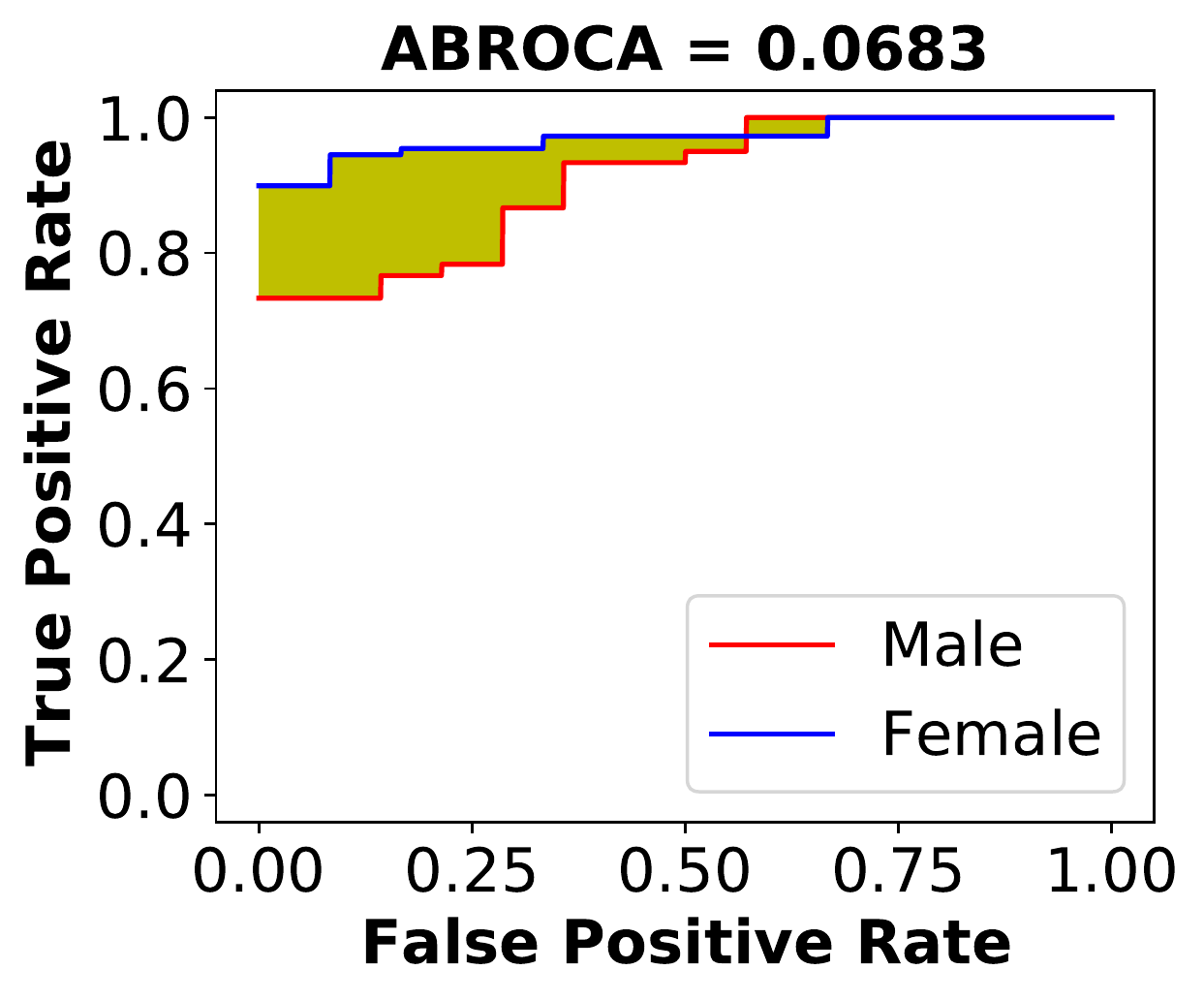}
    \caption{MLP}
\end{subfigure}
\bigskip
\vspace{-5pt}
\begin{subfigure}{.32\linewidth}
    \centering
    \includegraphics[width=\linewidth]{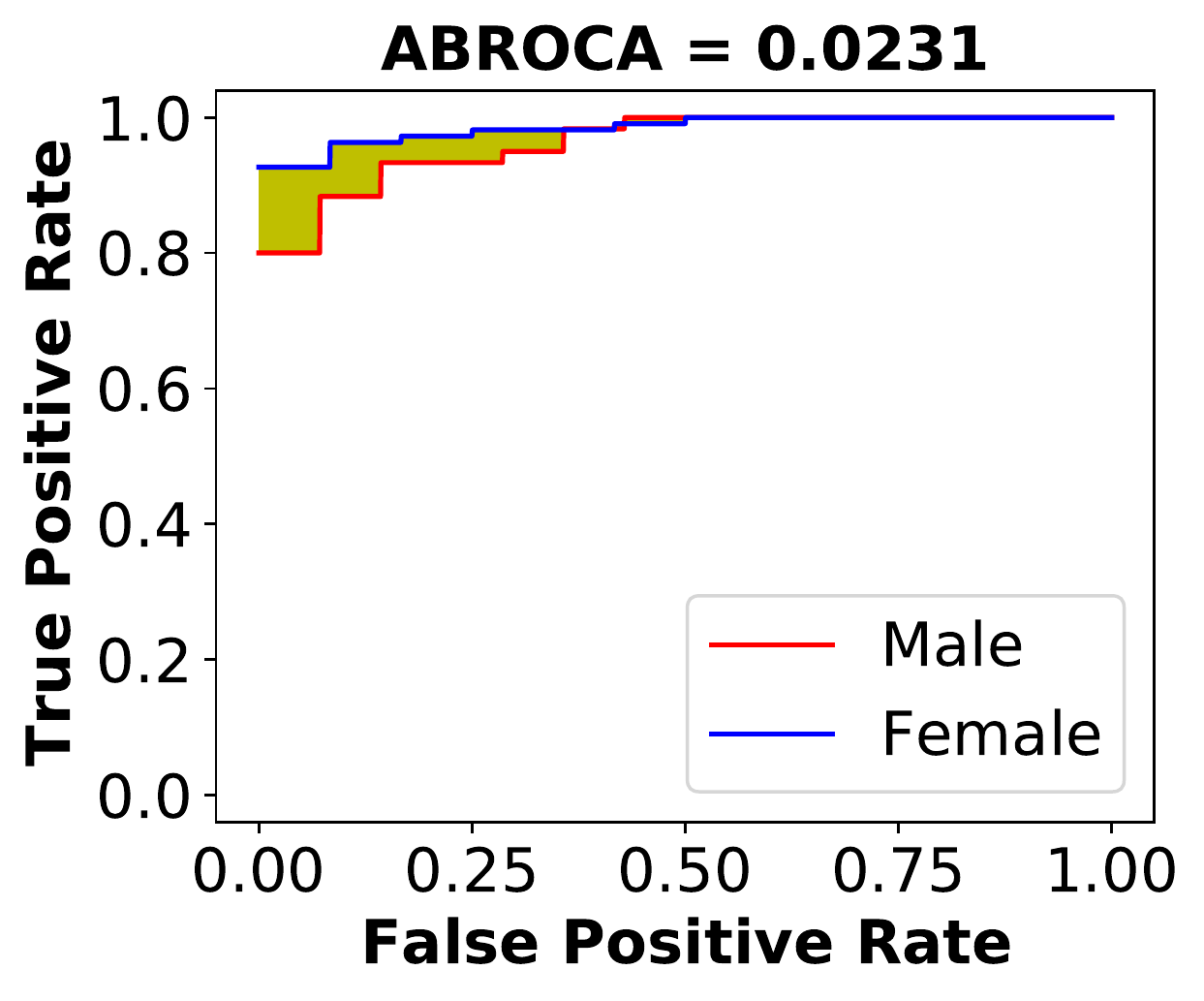}
    \caption{SVM}
\end{subfigure}
\hfill
\vspace{-5pt}
\begin{subfigure}{.32\linewidth}
    \centering
    \includegraphics[width=\linewidth]{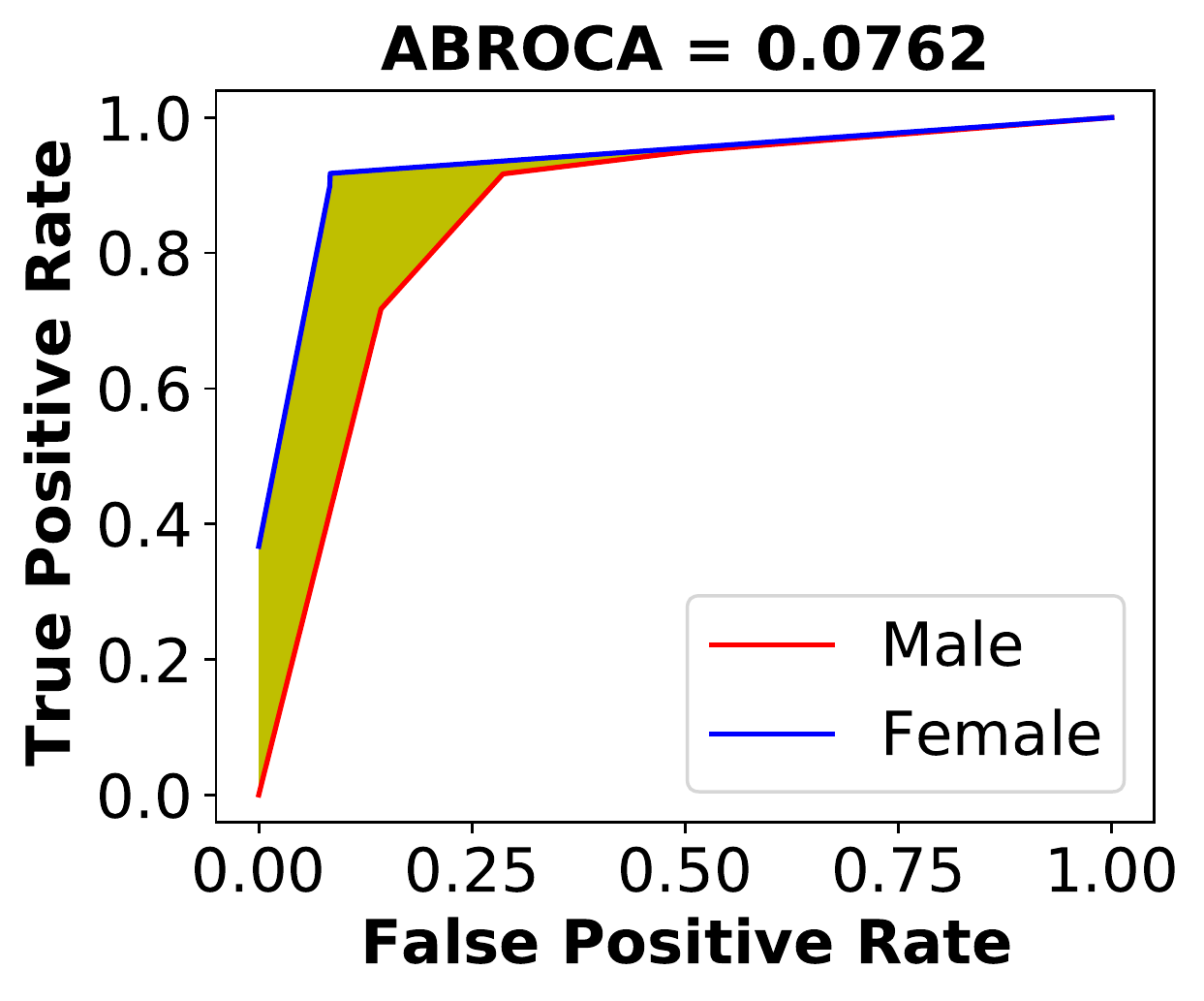}
    \caption{Agarwal's}
\end{subfigure}
\vspace{-5pt}
 \hfill
\begin{subfigure}{.32\linewidth}
    \centering
    \includegraphics[width=\linewidth]{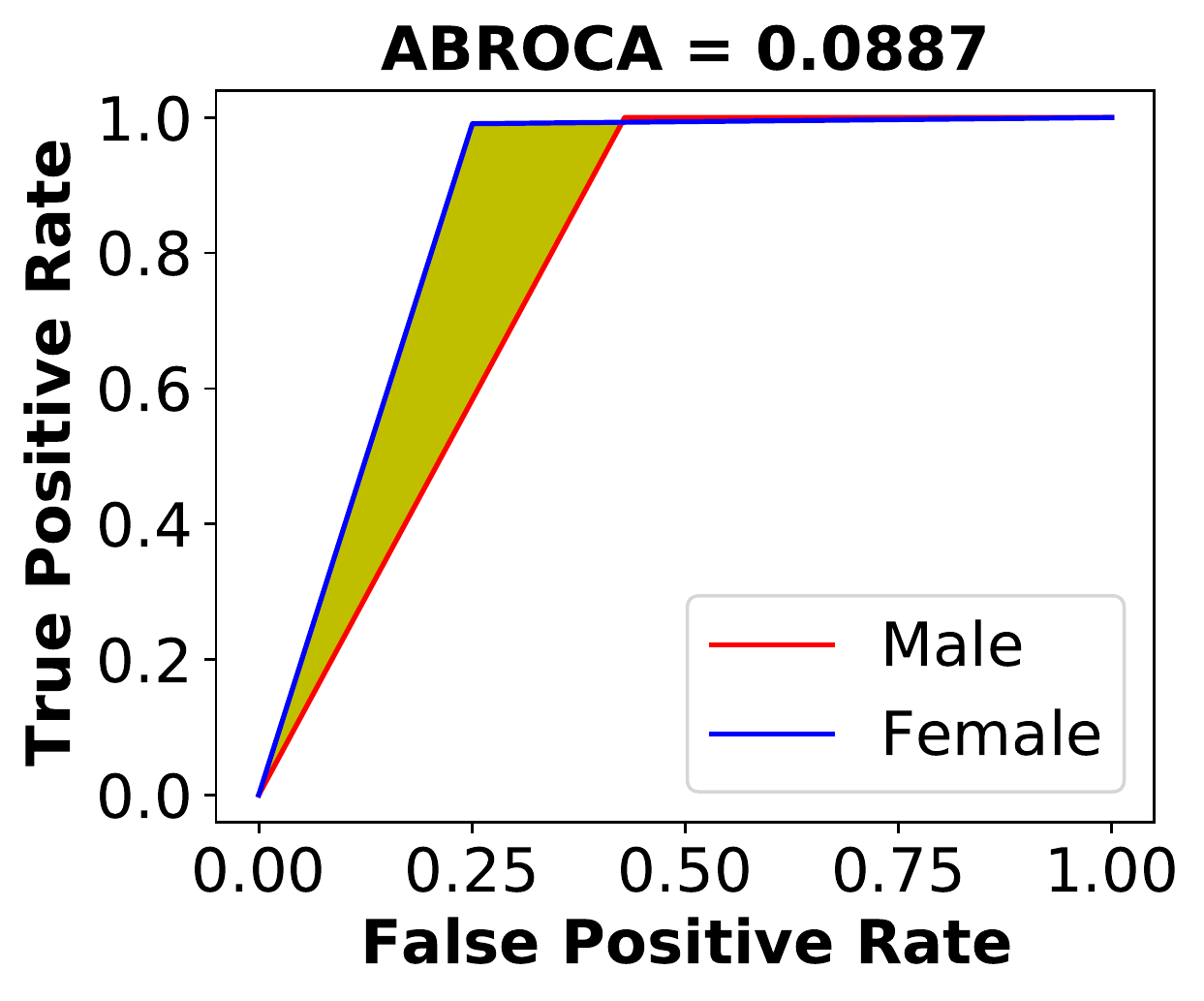}
    \caption{AdaFair}
\end{subfigure}
\caption{Student performance: ABROCA slice plots}
\vspace{0pt}
\label{fig:student_por_abroca}

\end{figure*}

\textbf{Student performance dataset}.
\label{subsubsec:student}
In general, all models show good accuracy (balanced accuracy) on predicting students' performance (Table \ref{tbl:student_performance_por}). MLP and AdaFair models fairly guarantee the fairness of results on most measures. Besides, the values of fairness measures also do not vary significantly across predictive models (Fig. \ref{fig:Variation_measures}-d), although the ABROCA slices are quite different in shape (Fig. \ref{fig:student_por_abroca}).
\begin{table}[!h]
\centering
\caption{Student performance: performance of predictive models}\label{tbl:student_performance_por}
\begin{tabular}{lcccccc}
\hline
\textbf{Measures} &  
\multicolumn{1}{c}{\textbf{\phantom{aa}DT\phantom{aa}}} & 
\multicolumn{1}{c}{\textbf{\phantom{aa}NB\phantom{aa}}} &
\multicolumn{1}{c}{\textbf{\phantom{aa}MLP\phantom{aa}}} & 
\multicolumn{1}{c}{\textbf{\phantom{a}SVM}\phantom{a}} &
\multicolumn{1}{c}{\textbf{\phantom{a}Agarwal's\phantom}} &
\multicolumn{1}{c}{\textbf{AdaFair}}\\ \hline
Accuracy            & 0.9333 & 0.8974 & 0.9077  & 0.9231 & 0.8923 & \textbf{0.9487}\\
Balanced accuracy   & \textbf{0.8639} & 0.8595 & 0.7840 & 0.7441 & 0.8565 & 0.8240\\
Statistical parity  &-0.0382 &-0.0509 &-0.0630  & \textbf{0.0151} & -0.0209 & -0.0255\\
Equal opportunity   & 0.0125 & 0.0174 & 0.03   & 0.0183 & 0.0176 & \textbf{0.0092} \\
Equalized odds      & 0.1316 & 0.2198 & \textbf{0.1252}  & 0.3279 & 0.2200 & 0.1877\\
Predictive parity   & \textbf{0.0456} & 0.0591 & 0.0601  & 0.0944 & 0.0577 & 0.0639\\
Predictive equality & 0.1190 & 0.2024 & \textbf{0.0952}  & 0.3095 & 0.2024 & 0.1786\\
Treatment equality  & 2.0    & 7.5    & \textbf{0.3333}  & 0.5    & 9.75 & \textbf{0.3333}\\
ABROCA              & 0.0575 & 0.0686 & 0.0683 & \textbf{0.0231} & 0.0762 & 0.0887\\
\hline
\end{tabular}
\vspace{-10pt}
\end{table}

\begin{figure*}[!h]
\centering
\vspace{-5pt}
\begin{subfigure}{.32\linewidth}
    \centering
    \includegraphics[width=\linewidth]{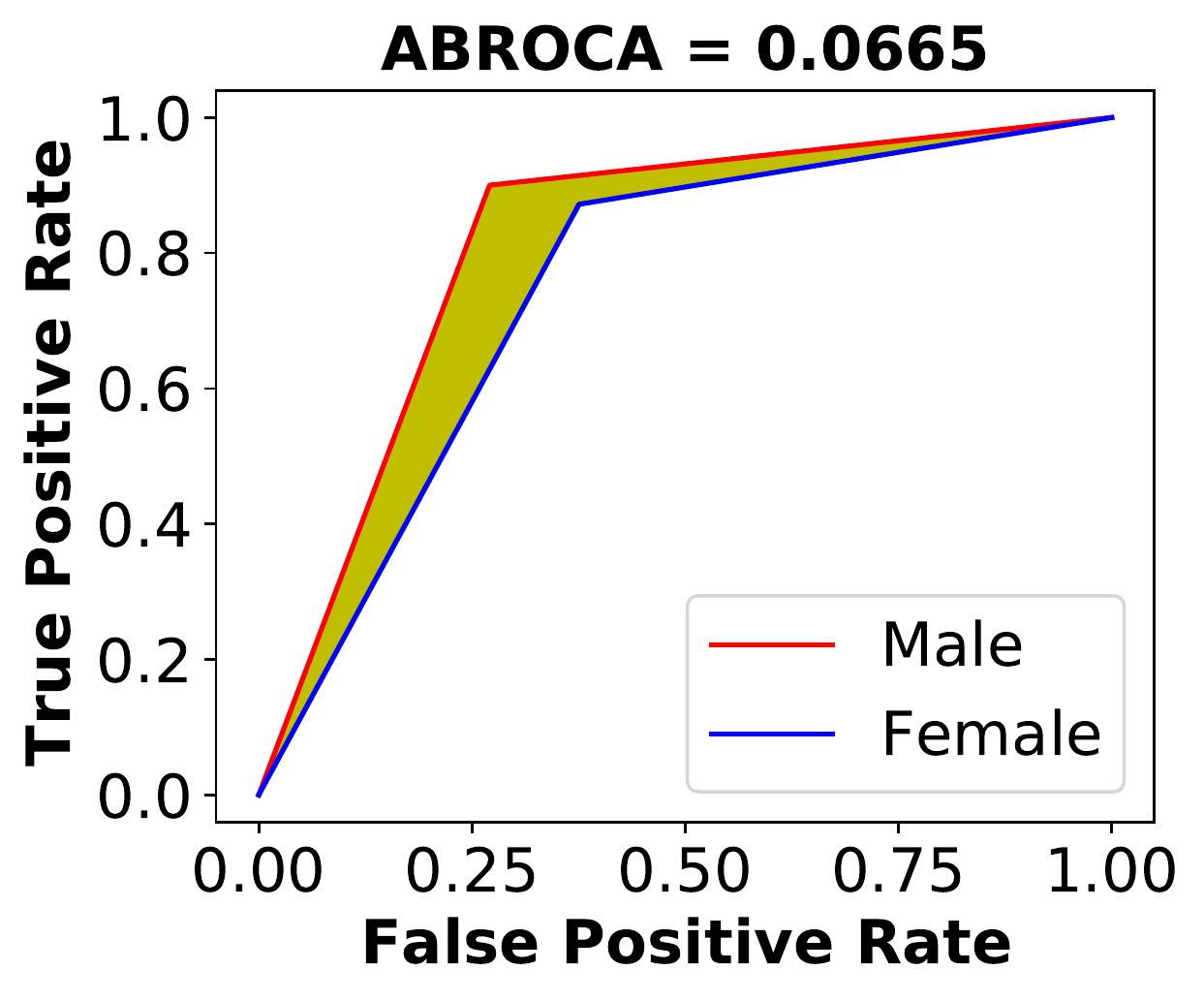}
    \caption{DT}
\end{subfigure}
\hfill
\vspace{-5pt}
\begin{subfigure}{.32\linewidth}
    \centering
    \includegraphics[width=\linewidth]{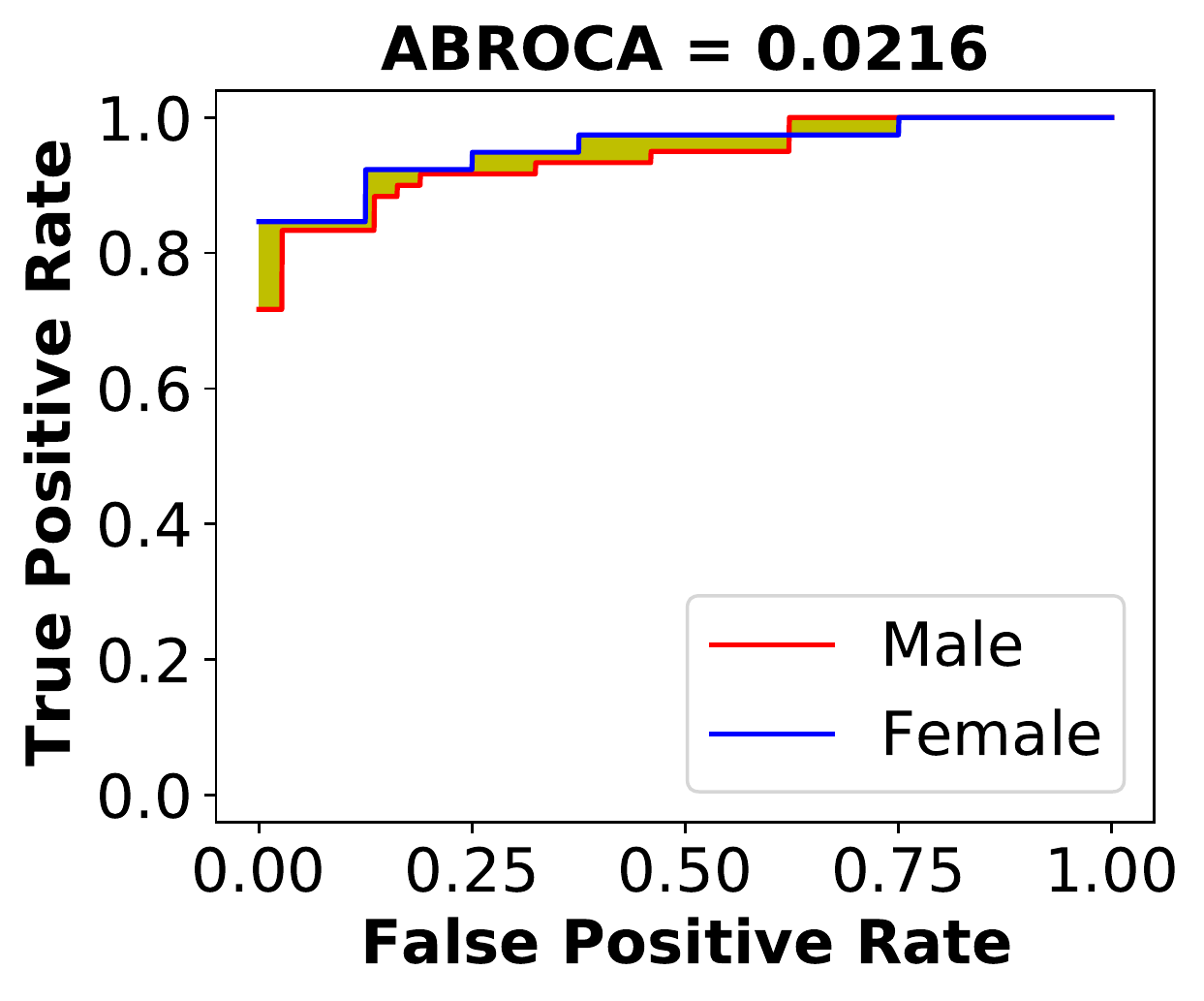}
    \caption{NB}
\end{subfigure}    
\hfill
\begin{subfigure}{.32\linewidth}
    \centering
    \includegraphics[width=\linewidth]{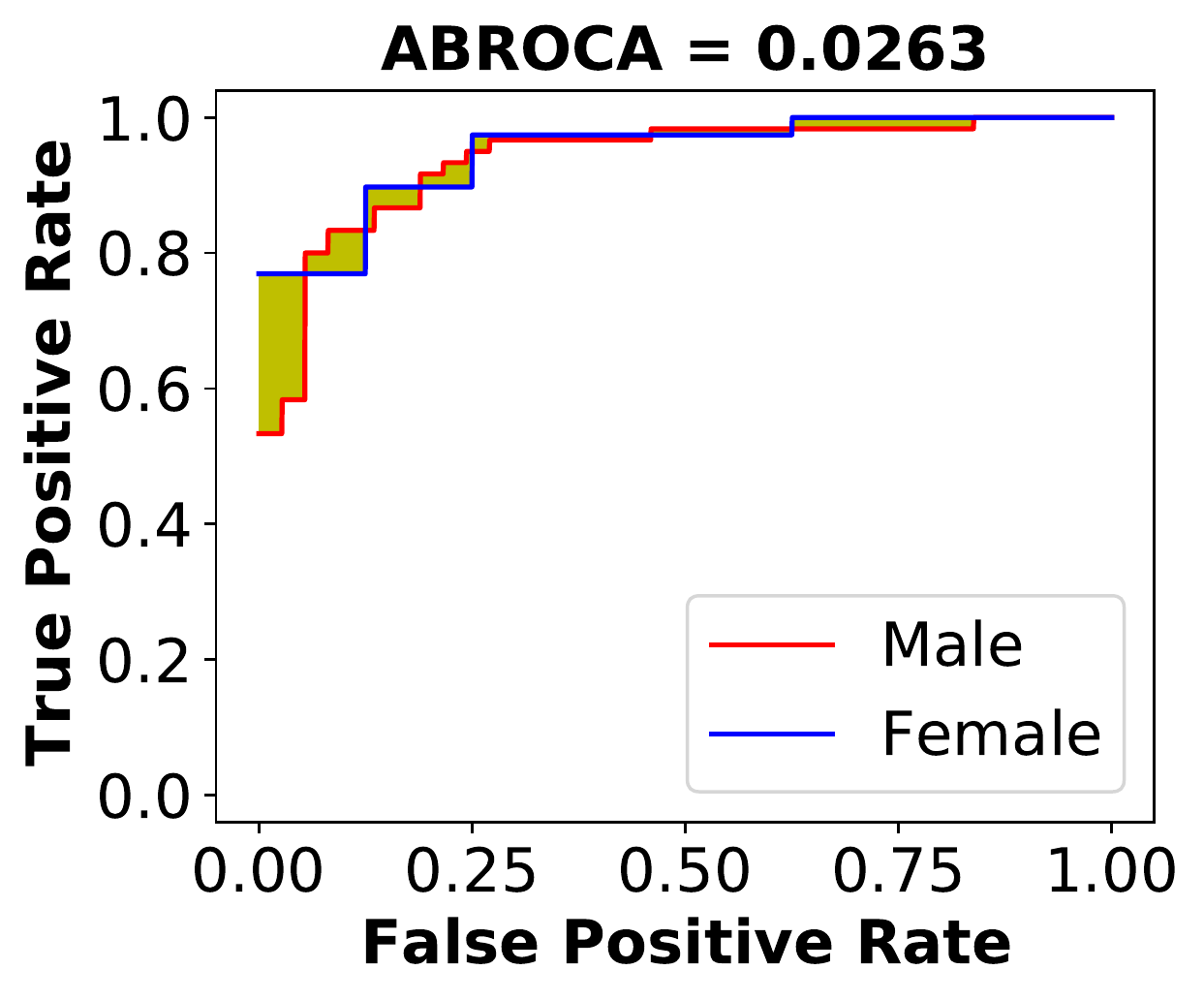}
    \caption{MLP}
\end{subfigure}
\bigskip
\vspace{-5pt}
\begin{subfigure}{.32\linewidth}
    \centering
    \includegraphics[width=\linewidth]{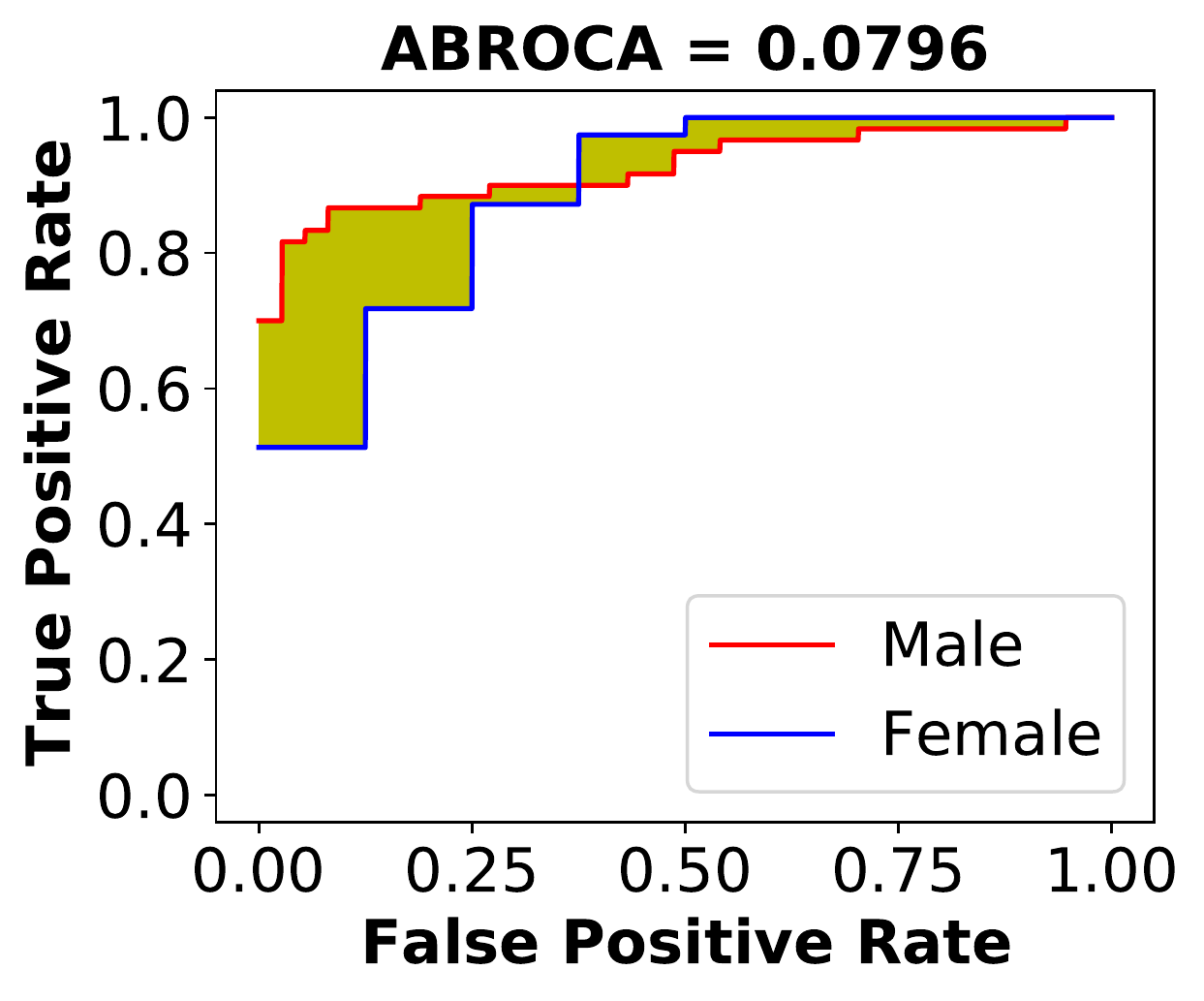}
    \caption{SVM}
\end{subfigure}
\hfill
\vspace{-5pt}
\begin{subfigure}{.32\linewidth}
    \centering
    \includegraphics[width=\linewidth]{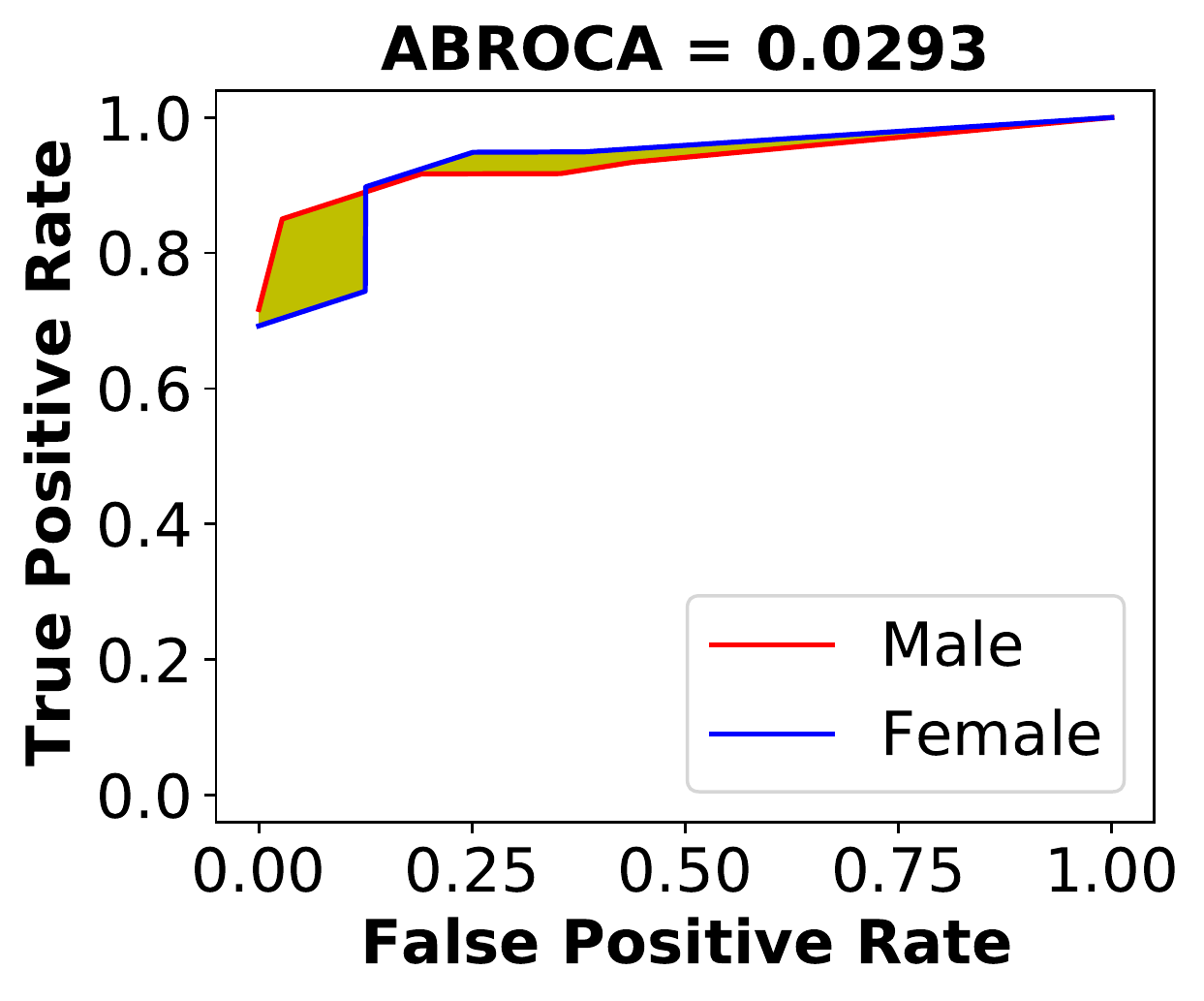}
    \caption{Agarwal's}
\end{subfigure}
\vspace{-5pt}
 \hfill
\begin{subfigure}{.32\linewidth}
    \centering
    \includegraphics[width=\linewidth]{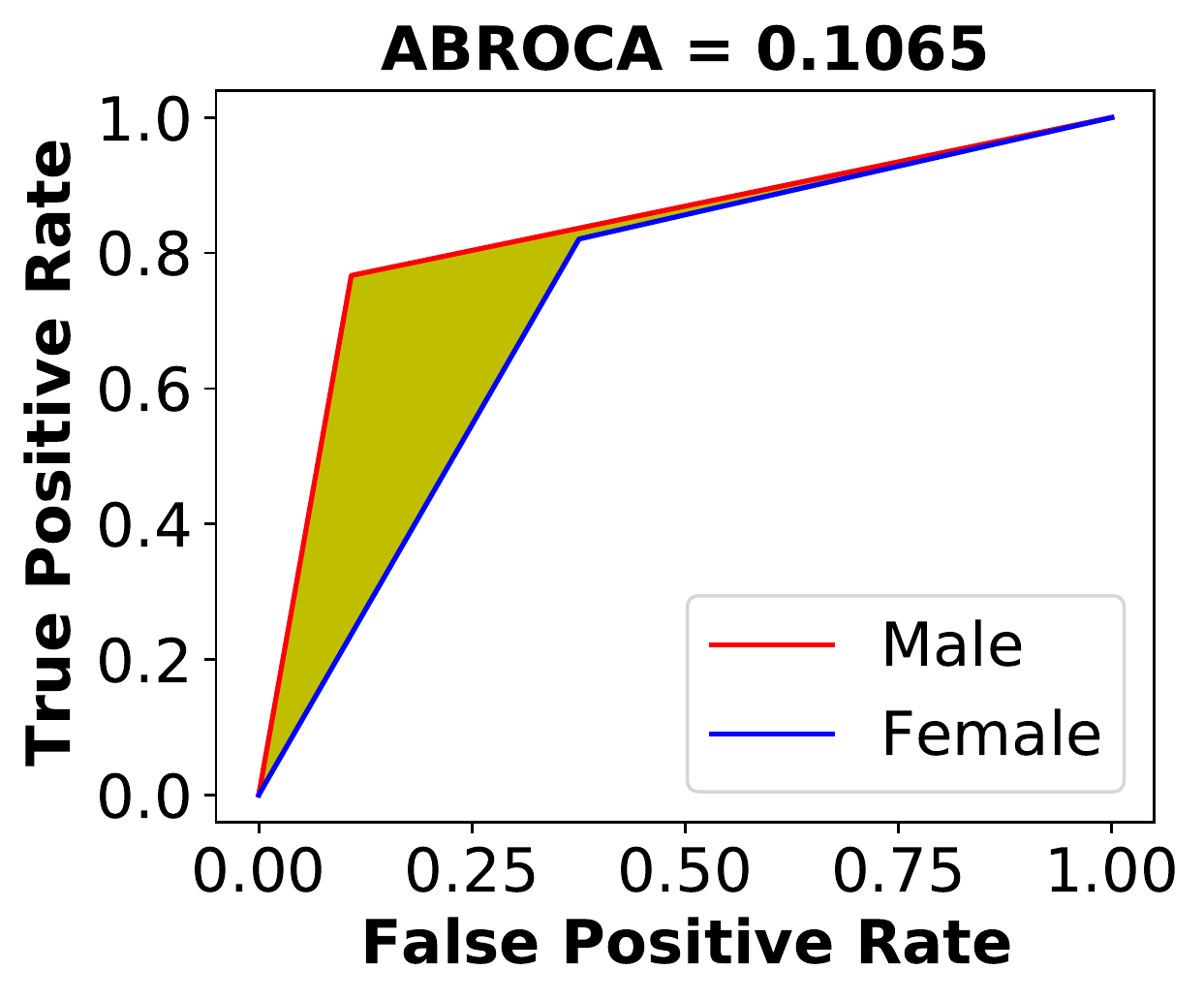}
    \caption{AdaFair}
\end{subfigure}
\caption{xAPI-Edu-Data: ABROCA slice plots.}
\vspace{-13pt}
\label{fig:xAPI-Edu-Data_abroca}
\end{figure*}

\textbf{xAPI-Edu-Data dataset} 
\label{subsubsec:xAPI-Edu-Data}
This is a surprising dataset because the traditional classification methods show a better performance not only in terms of accuracy/balanced accuracy measures but also w.r.t. fairness measures (Table \ref{tbl:xAPI-Edu-Data}). In addition, variation in the values of fairness measures across the predictive models is not significant, as shown in Fig. \ref{fig:Variation_measures}-e, except for the ABROCA measure with a noticeable change in the shape (Fig. \ref{fig:xAPI-Edu-Data_abroca}). 

\begin{table}[!h]
\centering
\caption{xAPI-Edu-Data: performance of predictive models}\label{tbl:xAPI-Edu-Data}
\begin{tabular}{lcccccc}
\hline
\textbf{Measures} &  
\multicolumn{1}{c}{\textbf{\phantom{aa}DT\phantom{aa}}} & 
\multicolumn{1}{c}{\textbf{\phantom{aa}NB\phantom{aa}}} &
\multicolumn{1}{c}{\textbf{\phantom{aa}MLP\phantom{aa}}} & 
\multicolumn{1}{c}{\textbf{\phantom{a}SVM}\phantom{a}} &
\multicolumn{1}{c}{\textbf{\phantom{a}Agarwal's\phantom}} &
\multicolumn{1}{c}{\textbf{AdaFair}}\\ \hline
Accuracy            & 0.8333 & 0.8750 & 0.8750 &  0.8611 & \textbf{0.8681} & 0.8056\\
Balanced accuracy   & 0.8    & \textbf{0.8970} & 0.8545 &  0.8505 & 0.8859 & 0.8162\\
Statistical parity  &\textbf{-0.1274} &-0.2608 &-0.2112  &-0.2209 & -0.2505 & -0.2292 \\
Equal opportunity   & 0.0282 & 0.0974 & 0.0654 & \textbf{0.0308}  & 0.0974 & 0.0538\\
Equalized odds      & 0.1329 & 0.1954 & \textbf{0.1262} & 0.2706 & 0.1684 & 0.3207\\
Predictive parity   & 0.0752 & 0.0074 & 0.0654 & 0.0088 & 0.0122 & \textbf{0.0057}\\
Predictive equality & 0.1047 & 0.0980 & \textbf{0.0608} & 0.2399 & 0.0709 & 0.2669\\
Treatment equality  & 1.0667 & -8.0   & \textbf{0.0}    &-0.2667 & -2.0 & -1.1667\\
ABROCA              & 0.0665 & \textbf{0.0216} & 0.0263 & 0.0796 & 0.0293 & 0.1065\\
\hline
\end{tabular}
\vspace{-7pt}
\end{table}


Regarding the \emph{treatment equality} measure, this measure is entirely different from all other measures with an extensive range of values, which is visualized in Fig. \ref{fig:Variation_measures}-f\footnote{We use the abbreviations of the fairness measures and datasets in Fig. \ref{fig:Variation_measures}}. On the \emph{PISA} datasets, this TE measure shows the best values across predicted models, followed by \emph{Law school} and \emph{Student Academics} datasets.

\textbf{Summary of results:} In general, \emph{ABOCA} is the measure with the lowest variability across predictive methods and datasets. It also clearly presents the ML model's accuracy variation over each value of the protected attribute. \emph{Equal opportunity} and \emph{predictive parity} also have a slight variation across methods and datasets. \emph{Equalized odds}, to some extent, can represent two measures \emph{
equal opportunity} and \emph{predictive equality} as it is the sum of the other two metrics. Furthermore, \emph{treatment equality} has a very wide range of values (sometimes the value may not be bounded), making it difficult to compare and evaluate.

\vspace{-5pt}
\subsection{Effect of varying grade threshold on fairness}
\label{subsec:threshold}
\begin{figure*}[!h]
\centering
\vspace{-5pt}
\begin{subfigure}{.32\linewidth}
    \centering
    \includegraphics[width=\linewidth]{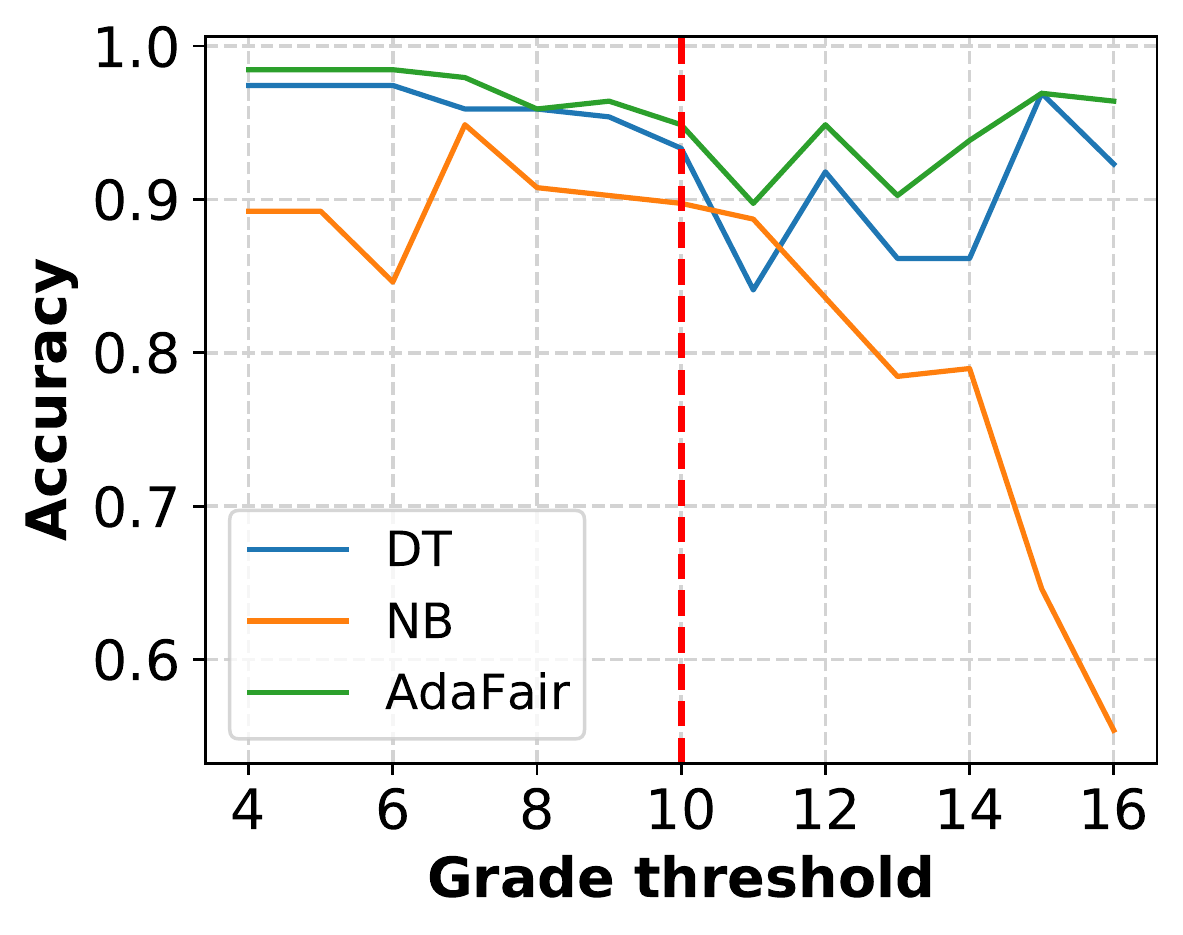}
    \caption{Accuracy}
\end{subfigure}
\hfill
\vspace{-5pt}
\begin{subfigure}{.32\linewidth}
    \centering
    \includegraphics[width=\linewidth]{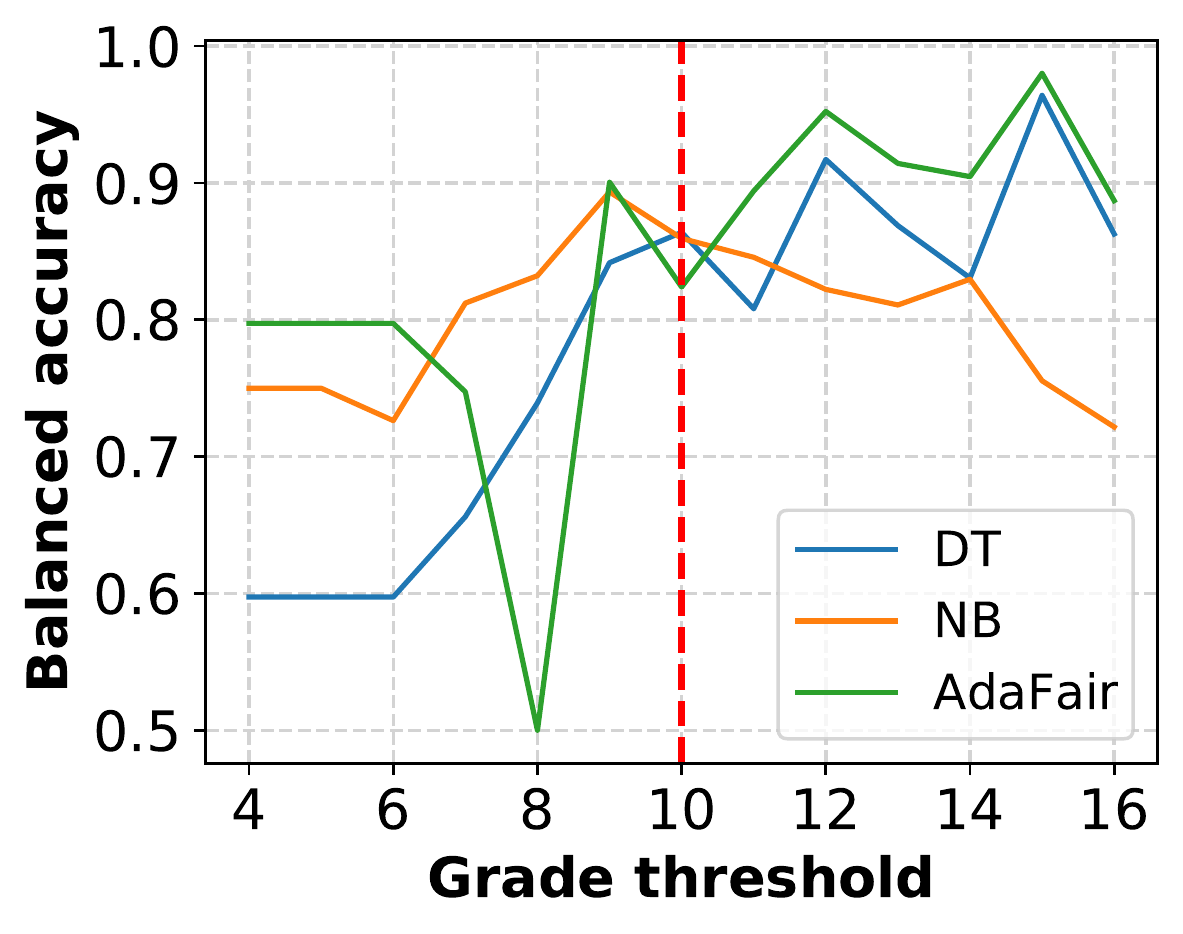}
    \caption{Balanced accuracy}
\end{subfigure}    
\hfill
\begin{subfigure}{.32\linewidth}
    \centering
    \includegraphics[width=\linewidth]{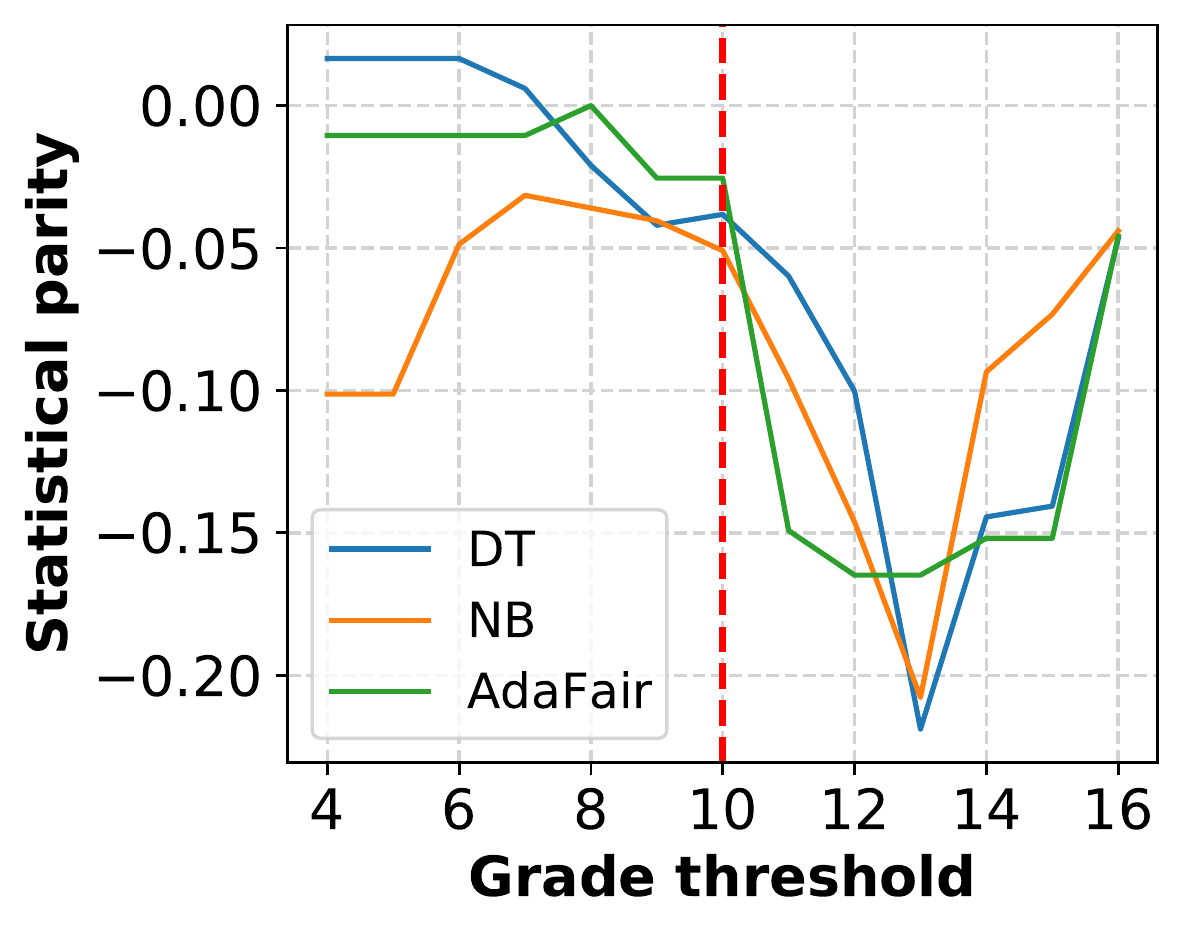}
    \caption{Statistical parity}
\end{subfigure}
\bigskip
\vspace{-5pt}
\begin{subfigure}{.32\linewidth}
    \centering
    \includegraphics[width=\linewidth]{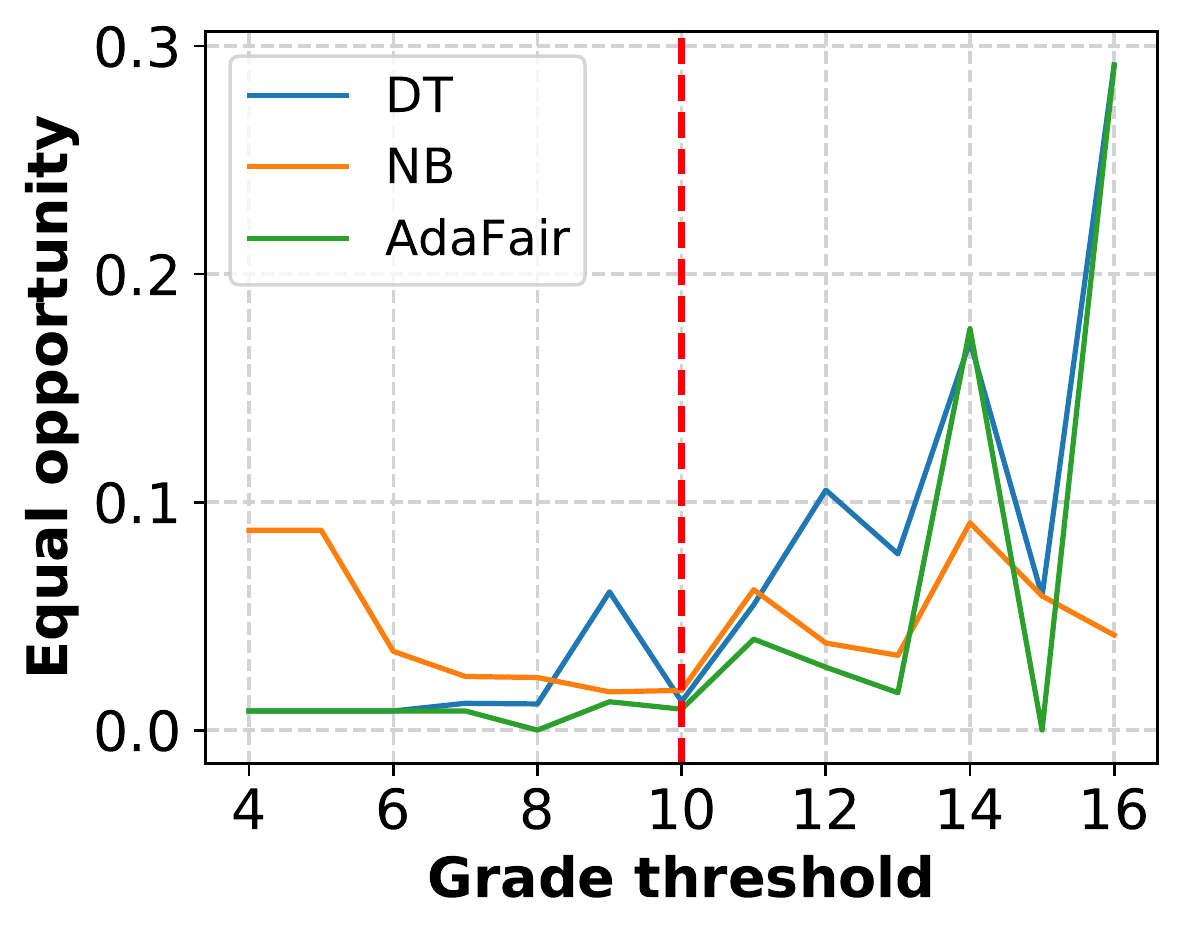}
    \caption{Equal opportunity}
\end{subfigure}
\hfill
\vspace{-5pt}
\begin{subfigure}{.32\linewidth}
    \centering
    \includegraphics[width=\linewidth]{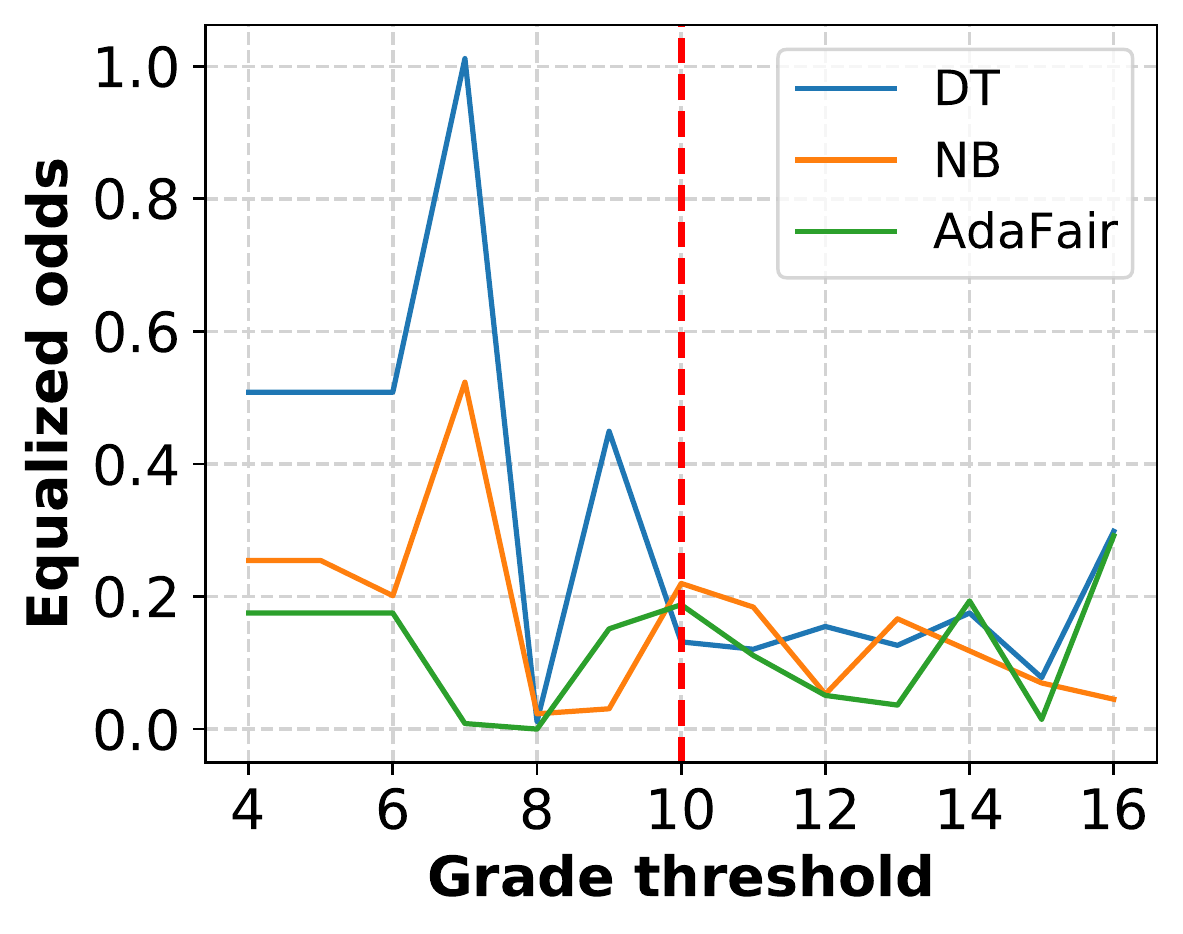}
    \caption{Equalized odds}
\end{subfigure}
\vspace{-5pt}
 \hfill
\begin{subfigure}{.32\linewidth}
    \centering
    \includegraphics[width=\linewidth]{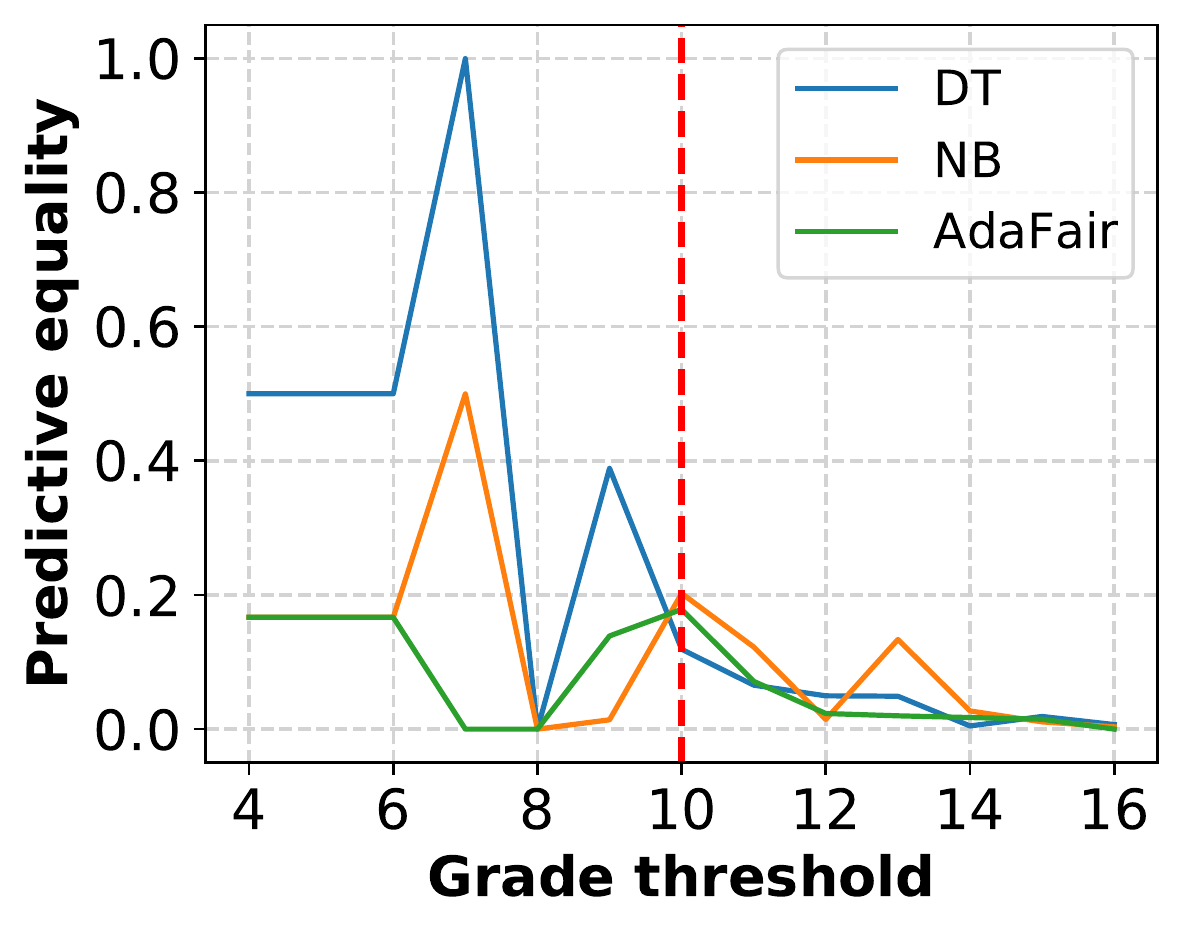}
    \caption{Predictive parity}
\end{subfigure}
\bigskip
\vspace{-5pt}
\begin{subfigure}{.32\linewidth}
    \centering
    \includegraphics[width=\linewidth]{Thres.Predictive_equality.pdf}
    \caption{Predictive equality}
\end{subfigure}
\hfill
\vspace{-5pt}
\begin{subfigure}{.32\linewidth}
    \centering
    \includegraphics[width=\linewidth]{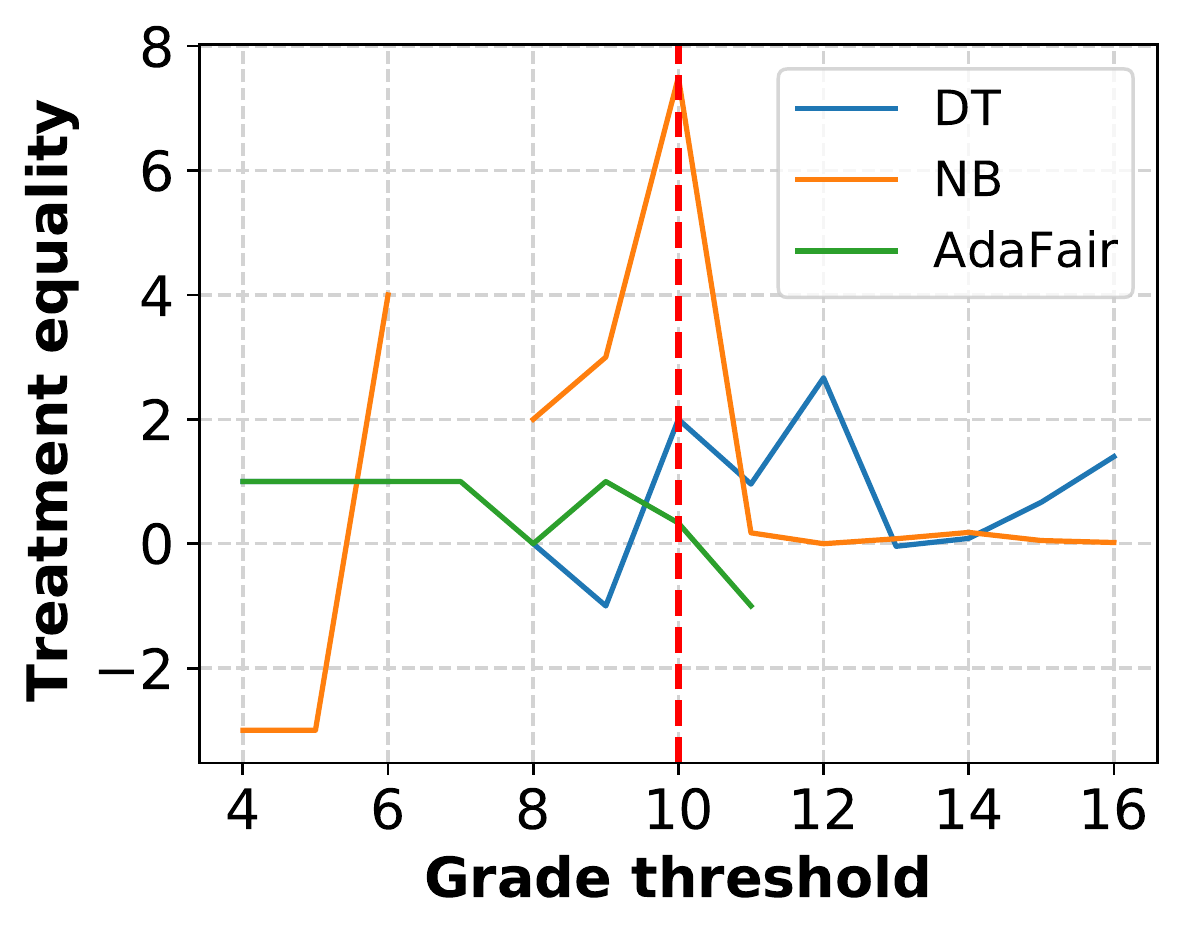}
    \caption{Treatment equality}
\end{subfigure}
\vspace{-5pt}
 \hfill
\begin{subfigure}{.32\linewidth}
    \centering
    \includegraphics[width=\linewidth]{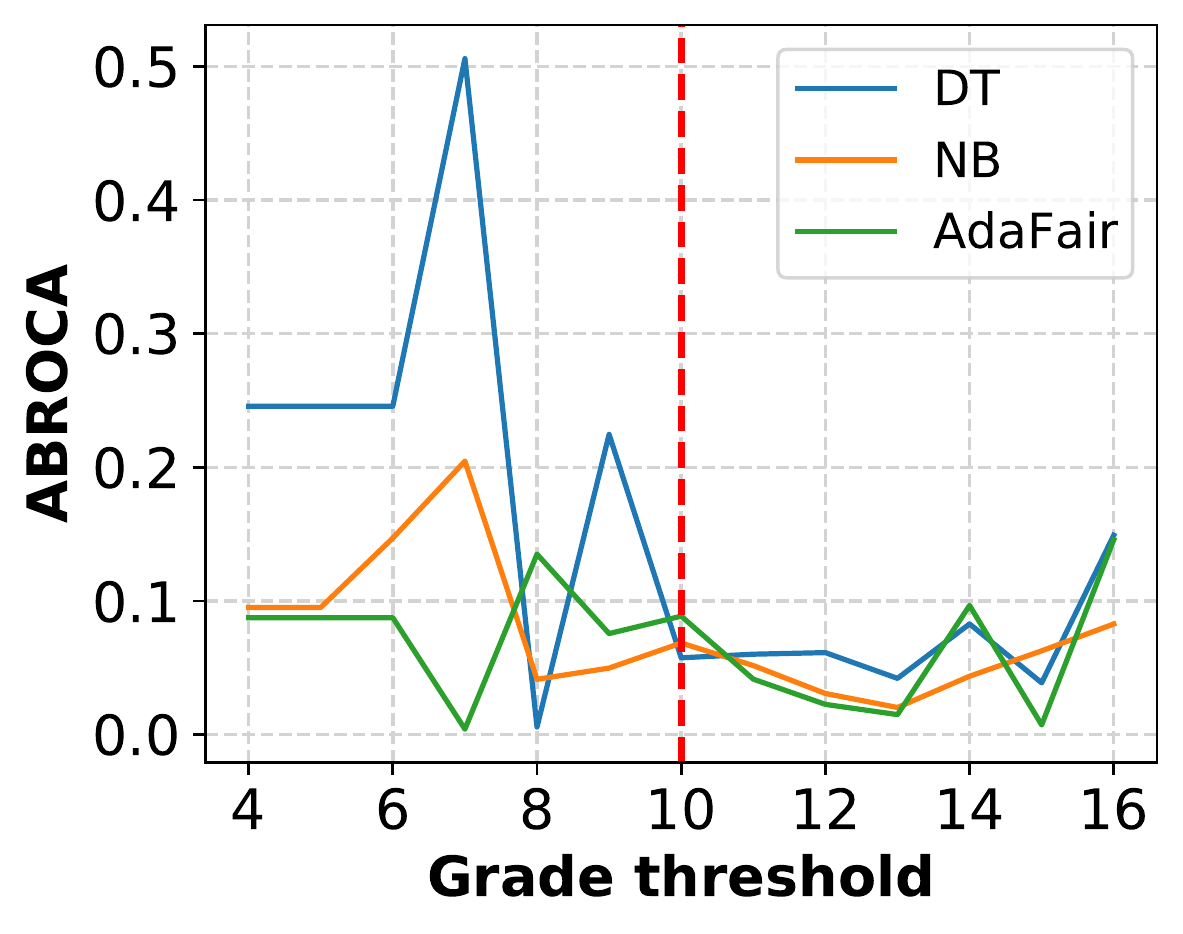}
    \caption{ABROCA}
\end{subfigure}
\caption{Accuracy and fairness interventions with varying grade threshold on Student performance dataset (Decision Tree method).}
\vspace{-10pt}
\label{fig:threshold}
\end{figure*}

Grade thresholds are often chosen as a basis for determining whether a candidate passes or fails an exam. In the student performance dataset, 10 (out of 20) is selected as the grade threshold \cite{cortez2008using,lequy2022survey}. However, the selection of a threshold can affect the fairness of the predictive models, as shown in the IPUMS Adult dataset \cite{ding2021retiring}. Hence, we investigate the effect of grade threshold on fairness by varying the threshold in a range of [4, 16], corresponding to 25\% to 75\% of the maximum grade (20). The results in Fig. \ref{fig:threshold} show that all fairness measures are affected by the grade threshold. When the grade threshold is gradually increased, the predictive models tend to be fairer (shown on the measures: equalized odds, predictive equality, and ABROCA). The opposite trend is observed in the remaining measures (except the treatment equality measure). Regarding the balanced accuracy, two models (DT and AdaFair) tend to predict more accurately. The NB model has a decreasing accuracy after the threshold is increased.

\vspace{-5pt}
\section{Conclusion and outlooks}
\label{sec:conclusion}
In this work, we evaluate seven popular group fairness measures for student performance prediction problems. We conduct experiments using four traditional ML models and two fairness-aware ML methods on five educational datasets. Our experiments reflect variations and correlations of fairness measures across datasets and predictive models. The results provide a overview picture for the selection of fairness measure in a specific case. Besides, we investigate the effect of varying grade thresholds on the accuracy and fairness of ML models. The preliminary results suggest that choosing the threshold is an important factor contributing to ensuring fairness in the output of the ML models. In the future, we plan to extend our evaluation of fairness w.r.t. more than one protected attribute, such as gender and race, and further explore the correlation between groups of fairness notions.
\vspace{-7pt}
\section*{Acknowledgments}
\vspace{-7pt}

The work of the first author is supported by the Ministry of Science and Culture of Lower Saxony, Germany, within the Ph.D. program ``LernMINT: Data-assisted teaching in the MINT subjects”. The work of the second author is funded by the German Research Foundation (DFG Grant NI-1760/1-1), project ``Managed Forgetting''.

%
%
%
\vspace{-5pt}
\bibliographystyle{splncs04}
\bibliography{a_bibliography}
\end{document}